\documentclass{article}

\usepackage[dandb, final]{neurips_2025}
\usepackage[utf8]{inputenc} % allow utf-8 input
\usepackage[T1]{fontenc}    % use 8-bit T1 fonts
\usepackage{hyperref}       % hyperlinks
\usepackage{url}            % simple URL typesetting
\usepackage{booktabs}       % professional-quality tables
\usepackage{amsfonts}       % blackboard math symbols
\usepackage{nicefrac}       % compact symbols for 1/2, etc.
\usepackage{microtype}      % microtypography
\usepackage{xcolor}         % colors

\usepackage{array}      % Enhanced table features
\usepackage{multirow}   % Merge rows in tables
\usepackage{multicol}   % Multiple column layout
\usepackage{makecell}   % Cell formatting in tables
\usepackage{tabularx}   % Table width control
\usepackage{booktabs}   % Professional quality tables
\usepackage{colortbl}   % Color support in tables

\usepackage{amssymb}    % Mathematical symbols
\usepackage{xcolor}     % Color support
\usepackage{pifont}     % Dingbats and symbols
\usepackage{wasysym}    % Additional symbols

\usepackage{caption}    % Caption customization
\usepackage{subcaption} % Sub-figure captions
\usepackage{float}      % Float placement control
\usepackage{adjustbox}  % Size adjustment for boxes
\usepackage{placeins}   % Float barrier control

\usepackage{hyperref}   % Hyperlinks and references
\usepackage{listings}   % Code listing and syntax highlighting
\usepackage{enumitem}   % Customized lists
\usepackage{lineno}

\newcommand{\cmark}{\ding{51}}    % Checkmark symbol
\newcommand{\xmark}{\ding{55}}    % Cross symbol
\newcommand{\yes}{\textcolor{green}{\cmark}}  % Green checkmark
\newcommand{\no}{\textcolor{red}{\xmark}}     % Red cross

\newcommand{\benchmark}{{\fontfamily{ppl}\selectfont MedChain}}

\title{\benchmark: Bridging the Gap Between LLM Agents and Clinical Practice with Interactive Sequence}

\author{Jie Liu$^{1}$,
	Wenxuan Wang$^{2}$,
	Zizhan Ma$^3$, Guolin Huang$^4$, SU Yihang$^3$, 
	\\[0.5mm] \textbf{Kao-Jung Chang}$^{5,6}$, \textbf{Wenting Chen}$^1$, \textbf{Haoliang Li}$^1$, \textbf{Linlin Shen}$^4$, \textbf{Michael Lyu}$^3$ \\[4mm]
	$^1$The City University of Hong Kong~ $^2$Renmin University of China \\ [0.5mm] $^3$The Chinese University of Hong Kong~
	$^4$Shenzhen University \\ [0.5mm] $^5$National Yang Ming Chiao Tung University
	$^6$Taipei Veterans General Hospital\\ [0.5mm]
	{\small Project:~\href{https://github.com/ljwztc/MedChain}{https://github.com/ljwztc/MedChain}}
}

\begin{document}

\maketitle

\begin{abstract}
  Clinical decision making (CDM) is a complex, dynamic process crucial to healthcare delivery, yet it remains a significant challenge for artificial intelligence systems. While Large Language Model (LLM)-based agents have been tested on general medical knowledge using licensing exams and knowledge question-answering tasks, their performance in the CDM in real-world scenarios is limited due to the lack of comprehensive benchmark that mirror actual medical practice. To address this gap, we present MedChain, a dataset of 12,163 clinical cases that covers five key stages of clinical workflow. MedChain distinguishes itself from existing benchmarks with three key features of real-world clinical practice: personalization, interactivity, and sequentiality. Further, to tackle real-world CDM challenges, we also propose MedChain-Agent, an AI system that integrates a feedback mechanism and a MedCase-RAG module to learn from previous cases and adapt its responses. MedChain-Agent demonstrates remarkable adaptability in gathering information dynamically and handling sequential clinical tasks, significantly outperforming existing approaches. The relevant dataset and code will be released upon acceptance of this paper. 
\end{abstract}

\section{Introduction}
At the intersection of artificial intelligence and healthcare lies one of medicine's most complex challenges: Clinical Decision Making (CDM). In healthcare delivery, CDM demands not only the integration of diverse data sources and continuous assessment of evolving clinical scenarios, but also evidence-based judgments for diagnosis and treatment \cite{sutton2020overview}. While crucial for optimal patient care, this intricate process imposes significant cognitive demands on healthcare professionals, making it an ideal candidate for AI assistance \cite{sendak2020human}.

\begin{figure*}[t]
	\centering
	\includegraphics[width=\textwidth]{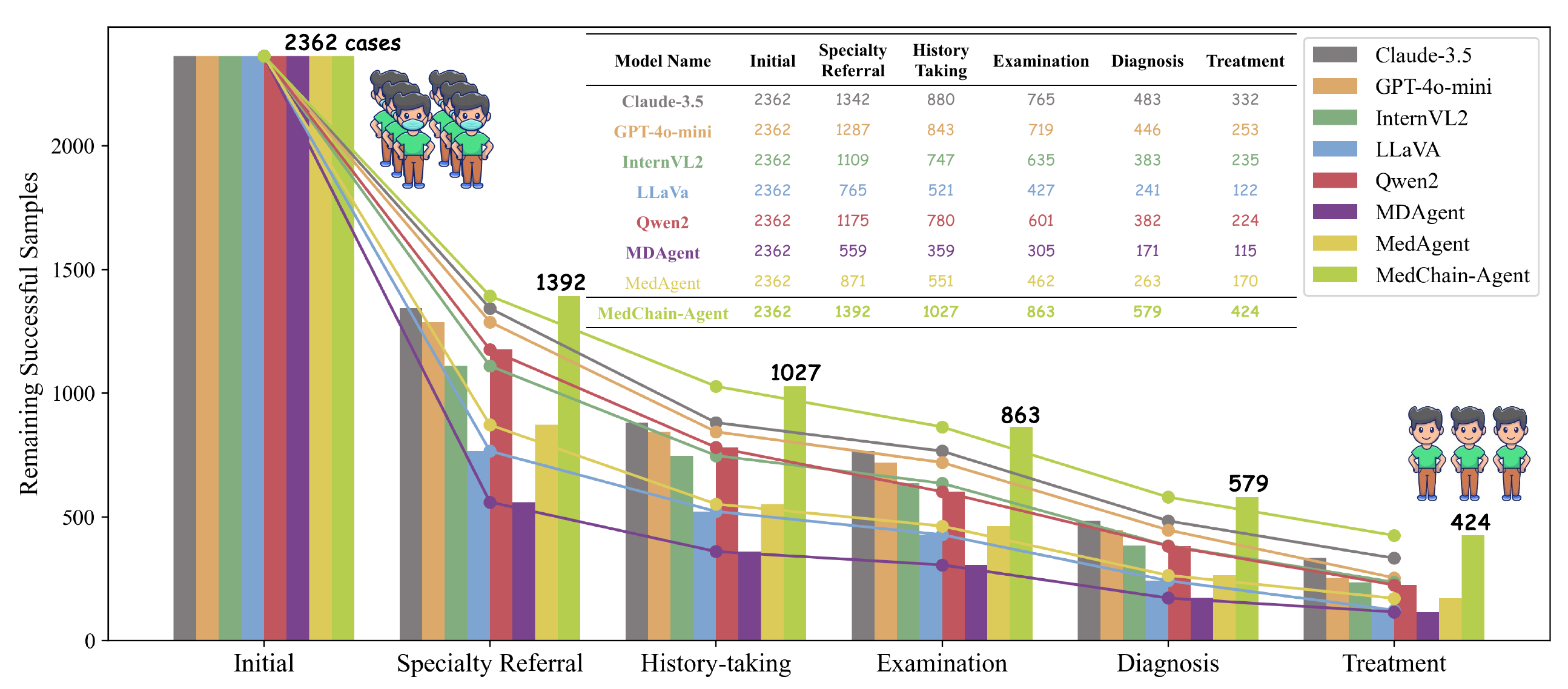}	
	\caption{\textbf{Demonstration of error propagation of CDM in \benchmark.} Starting with 2,362 initial cases, the diagram illustrates how diagnostic errors cascade through five clinical stages. Cases with incorrect diagnoses carry forward problematic information to subsequent stages, leading to a cumulative decrease in accuracy. After completing the treatment phase, we count cases that maintain correctness through each consecutive phase up to the each stage. Our \benchmark-Agent achieves best performance in CDM comparing with other SOTA methods.}
	\label{fig:sequential_demonstration}
\end{figure*}

Recent advances in Large Language Model (LLM)-based agents \cite{openai2023gpt4v,team2023gemini,gu2023don,shinn2024reflexion,guan2023leveraging,zhuangtoolchain} have emerged as an effective solution for complex decision-making tasks, from software development \cite{qian2024chatdev} to office automation \cite{wang2024officebench}. In the medical domain, these LLMs have demonstrated impressive performance on medical licensing exams \cite{singhal2023large,pal2022medmcqa} and knowledge-based assessments \cite{gilson2023does,eriksen2023use,liu2025comprehensive}. While LLMs have consistently scored well above passing thresholds in these benchmark \cite{singhal2023large}, it is crucial to recognize that these assessments fall short of capturing the complexity of real-world CDM, where errors can cascade through multiple decision stages, as illustrated in \autoref{fig:sequential_demonstration}. Based on our analysis, CDM exhibits three key characteristics.

\textbf{Firstly}, these benchmarks rarely account for patient-specific information such as past medical history and present illness \cite{pal2022medmcqa}, which significantly influence clinical decisions in real clinical scenarios. This omission fails to capture the nuanced context that often shapes \textit{personalized} diagnosis. \textbf{Secondly}, unlike real clinical scenarios where decisions build upon previous steps, existing benchmarks present clinical tasks as independent problems \cite{schmidgall2024agentclinic}, missing the critical interdependencies in the diagnostic process. In reality, clinical decision-making is a \textit{sequential} process where each step is contingent upon the preceding ones, and an error in one stage can profoundly impact subsequent decisions. \textbf{Thirdly}, most benchmarks present all relevant information upfront, providing a static, and comprehensive dataset \cite{tu2024towards}. However, real clinical workflow demand multiple rounds of dynamic information gathering through ongoing patient \textit{interaction}.

\noindent \textbf{\benchmark:} To address these critical gaps, we introduce \benchmark, a novel benchmark designed to evaluate LLM-based agents in real-world clinical scenarios. Specifically, \benchmark~comprises 12,163 diverse cases spanning 19 medical specialties and 156 sub-categories, including 7,338 medical images with corresponding reports. Each case progresses through five crucial stages: specialty referral, history-taking, examination, diagnosis, and treatment. Unlike existing benchmarks, \benchmark~uniquely emphasizes three key features. 1) \textit{Personalization:} Each case incorporates detailed patient-specific information. At first, agents are provided with only the patient's chief complaint and basic information. 2) \textit{Interactivity:} Information must be actively gathered through dynamic consultation from patient. 3) \textit{Sequentiality:} Decisions at each stage influence subsequent steps. Only after agent sequentially completes all five stages, the overall diagnostic process is evaluated.

\noindent \textbf{MedChain-Agent:} Given the novel features and challenges presented by this benchmark, existing agent frameworks struggle to address these issues adequately. To overcome these limitations, we propose MedChain-Agent, a multi-agent collaborative framework that enables LLMs with feedback mechanism and MedCase-RAG to dynamically gather information and handle sequential clinical tasks. Specifically, \benchmark-Agent facilitates a synergistic interplay among three specialized agent types: General Agents for task-specific expertise, a Summarizing Agent for insight synthesis, and a Feedback Agent for iterative refinement. This multi-layered, iterative approach ensures decisions are products of thorough analysis and diverse perspectives. Additionally, to address the multifaceted nature of CDM, which demands the integration of evidence-based research, and patient-specific factors, we incorporate a novel MedCase-RAG module into our MedChain-Agent framework. Unlike conventional medical RAG methods, MedCase-RAG dynamically expand its database and employs a structured approach to data representation, mapping each medical case into a 12-dimensional feature vector. This system enables efficient retrieval of relevant cases and helps the model make informed decisions.
%\wenxuan{The RAG introduced here to me is blunt, without any demand and motivation. We can add some motivations.} 

Our contributions are summarized as follows:

\begin{itemize}
	\item We represent the first effort to propose a CDM benchmark, \benchmark, providing a holistic assessment of diagnostic capabilities of LLLM-based agents, closely reflecting real-world patient care.
	\item We propose a multi-agent framework based on the characteristics of CDM, called MedChain-Agent. This system enables efficient retrieval of relevant cases and helps the model make informed decisions.
	\item Through extensive experiments, we compare the performance of existing works on MedChain and the superiority of MedChain-Agent in CDM and realibility.
\end{itemize}

\section{Related Works}

\subsection{Evaluation of LLM in Medcine}

Benchmarking plays a vital role as a key performance indicator, directing model improvements, pinpointing weaknesses, and shaping the course of model evolution. The evaluation of LLMs in medicine has primarily focused on testing general medical knowledge through structured assessments~\cite{ma2025beyond}. Leading benchmarks such as MultiMedQA \cite{singhal2023large} integrate various medical QA datasets (e.g., MedQA \cite{jin2021disease}, MedMCQA \cite{pal2022medmcqa}), emphasizing performance on medical licensing examination materials. Other benchmarks like PubMedQA \cite{jin2019pubmedqa} focus on research-oriented queries, while several Chinese medical benchmarks \cite{wang2023cmb,cai2024medbench} evaluate models through multiple-choice questions from medical licensing exams. While these benchmarks effectively assess general medical knowledge, they fail to capture three critical aspects of real-world clinical decision-making (see Appendix \autoref{tab:related_works}), i.e., personalization in patient care, the interactive nature of clinical consultations, and the sequential dependency of medical decisions, where each step builds upon previous findings.

Recently, several benchmarks \cite{wu2025towards,ouyang2024climedbench,liu2025asclepius,liang2024wsi,li2025towards,ding2025eagle} have been proposed to evaluate LLMs across diverse clinical scenarios and tasks, including information extraction, text summarization, and clinical outcome prediction. However, these benchmarks primarily consist of independent question-answer pairs, where each task is evaluated in isolation. Sequential decision-making is critical in medical practice, as each patient experiences a continuous journey from initial triage through treatment to recovery. Therefore, it is essential to evaluate how LLMs perform throughout this entire clinical pipeline. Our work distinguishes itself from existing benchmarks by focusing on sequential decision-making within interactive environments, specifically evaluating LLM performance in realistic clinical scenarios that require executing the complete patient care workflow.

\subsection{LLM-based Agent in Medicine}
LLM-based agents have demonstrated significant potential across various medical applications~\cite{wang-etal-2025-survey,wsiagent}, encompassing tasks such as medical examination questions, clinical diagnoses, and treatment plans. Recent research has explored different approaches: Agent Hospital \cite{agenthospitall} provides medical scenario simulation, while several frameworks \cite{clinicalagent,MEDAGENTS,mdagent} focus on specific medical stages with multi-agent architectures. Some works target specialized aspects, such as CoD \cite{chen2024cod} for interpretable diagnostics and Ehragent \cite{shi2024ehragent} for electronic health records (EHRs) analysis. Others, like Almanac Copilot \cite{zakka2024almanac}, assist clinicians with EMR-specific tasks. AI Hospital \cite{fan2024ai} explores interactive clinical scenarios, but it falls short in handling multi-modal medical imaging and lacks a comprehensive benchmark for evaluating multi-agent performance. To enhance these agents' capabilities, researchers have integrated Retrieval-Augmented Generation (RAG), as demonstrated by MIRAGE \cite{MIRAGE}'s search-enhanced framework and Medical Graph RAG \cite{GraphRAG}'s knowledge-based approach.

However, current approaches face two major limitations. First, existing frameworks focus on isolated medical tasks rather than providing seamless integration across different clinical stages, making them insufficient for complex scenarios requiring effective inter-stage communication \cite{coorapation}. Second, current medical RAG systems' reliance on chunk-based indexing leads to context inconsistency and computational inefficiencies \cite{RAGsurvey1, RAGsurvey2}, highlighting the need for more sophisticated approaches to medical knowledge integration.

\begin{figure*}[h]
	\centering
	\includegraphics[width=0.95\textwidth]{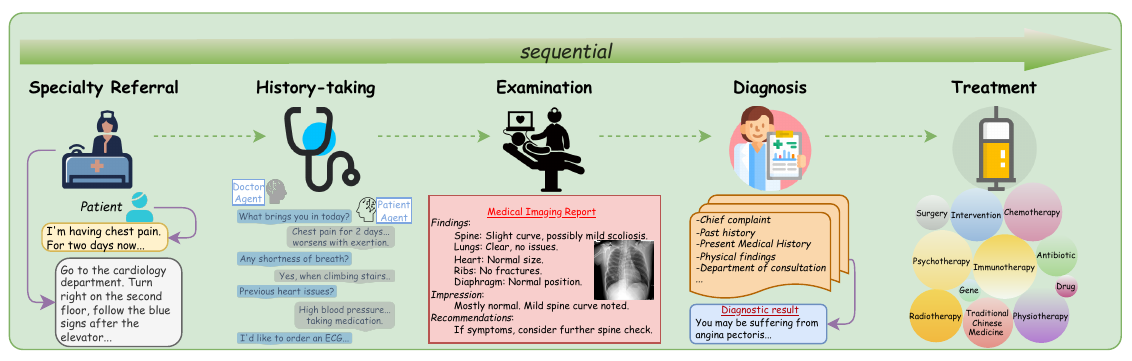}
	\caption{\textbf{\benchmark~Pipeline.} The \benchmark~is composed of a sequential medical process, including specialty referral, history-taking , examination, diagnosis, and treatment.}
	\label{fig:benchmark_overview}
\end{figure*}

\section{MedChain Benchmark}
\noindent \textbf{Overview.} We introduce MedChain, a comprehensive clinical decision-making benchmark designed to simulate real-world scenarios. Built upon 12,163 Electronic Health Records (EHRs) spanning 19 medical specialties and 156 sub-categories, including 7,338 medical images with reports, MedChain uniquely emphasizes three key characteristics: 
\begin{itemize}
	\item \textbf{\textit{Personalization}}: Each case includes detailed patient profiles that influence decision-making
	\item \textbf{\textit{Sequentiality}}: Cases involve multiple interconnected decision-making stages
	\item \textbf{\textit{Interactivity}}: Information must be actively gathered through dynamic consultation
\end{itemize}

\subsection{Data Collection}
\noindent \textbf{Data Source and Processing.} Our dataset is sourced from the Chinese medical website ``iiYi'' \footnote{https://www.iiyi.com}, which provides over 20,000 validated clinical cases spanning 28 disease categories. 
These cases are verified by professional doctors and have undergone de-identification to ensure patient privacy. We obtained formal permission from the website administrators to use the data for scientific research purposes. Each case typically contains the patient's chief complaint, medical history, examination results, treatment process, and other relevant information, which insure the \textbf{{personalization}} of \benchmark. 
Following the government standards \footnote{\hyperlink{http://www.nhc.gov.cn/caiwusi/s7785t/202309/914aec9618944ee2b36621d33517e576/files/f15b986edd1f4bf08bab3b1b8ce64d9a.pdf}{National Standards for Medical Items}} and Medical Subject Headings \footnote{\hyperlink{https://www.ncbi.nlm.nih.gov/mesh/1000048}{https://www.ncbi.nlm.nih.gov/mesh/1000048}}, we extracted and organized key information including patient basics, chief complaints, specialty referrals, examinations, imaging reports, diagnoses, and treatments. Cases with incomplete information were removed, resulting in 12,163 high-quality cases.

\noindent \textbf{Quality Control.} To ensure the highest standards of data integrity and clinical relevance in our benchmark, we implemented a rigorous quality control process involving a panel of five senior physicians, each with over 10 years of clinical experience. Our evaluation process examined a random sample of 6,000 cases (49.3\% of the dataset). We developed a standardized scoring system that presents physicians with comprehensive case information alongside six binary quality dimensions: disease prevalence, clinical relevance, accuracy of patient history, appropriateness of diagnostic procedures, correctness of diagnosis, and suitability of treatment recommendations. Physicians evaluate each dimension through yes/no responses, with cases satisfying all dimensions considered valid.

The quality assessment yielded strong results, with 94.7\% of evaluated cases meeting or exceeding our quality thresholds. Dimension-specific quality rates ranged from 92.9\% to 97.2\%, demonstrating consistently high standards across all evaluation criteria. Inter-rater reliability analysis produced a Cohen's kappa coefficient of 0.82, indicating substantial agreement among our expert reviewers. Cases that failed to meet the thresholds (5.3\%) were either revised or excluded from the final dataset to maintain benchmark integrity.

\subsection{Clinical Workflow Simulation}
\noindent \textbf{Sequential Stages.} MedChain simulates the complete clinical workflow, comprising five sequential tasks, each representing a different stage of the clinical decision-making process, as shown in \autoref{fig:benchmark_overview}. The results from each stage serve as inputs for the subsequent stage, creating a dependency where later decisions are influenced by the quality of earlier ones. This design guarantees the \textbf{{sequentiality}} of \benchmark, mimicking the interconnected nature of real-world clinical decision-making processes. The pipeline consists of: \textit{1) Specialty Referral:} Assessment of case urgency and appropriate department selection; \textit{2) History-taking:} Dynamic information gathering through doctor-patient dialogue; \textit{3) Examination:} Medical image analysis and report generation; \textit{4) Diagnosis:} Comprehensive diagnosis based on accumulated information; \textit{5) Treatment:} Treatment plan formulation considering patient-specific factors.

The construction pipeline of \benchmark~ is sketched out here. The standardization process and the format of each pipeline can be refer to \autoref{sec:append_benchmark_construction}. We demonstrate an example in Appendix \autoref{fig:case_show} and \autoref{fig:case_show_chinese}.

\begin{figure*}[t]
	\centering
	\includegraphics[width=0.95\textwidth]{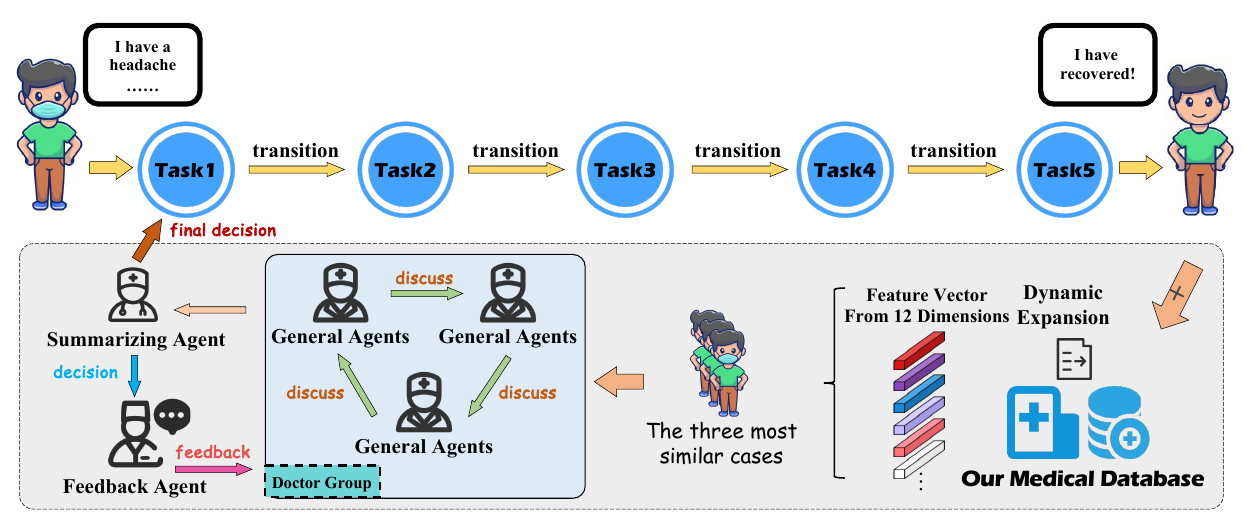}
	\caption{\textbf{MedChain-Agent framework.} Depicts a cyclical feedback medical multi-task system, where decisions are supported by retrieving similar past cases from a medical database.}
	\label{fig:agent_overview}
\end{figure*}

\subsection{Interaction Environment}
To simulate authentic doctor-patient consultations, we developed an \textbf{interactive} environment where LLM-based agents must actively gather information through dynamic interactions. We employ Gemma2 (9b) \cite{team2024gemma} as the patient agent, initializing it with pre-defined case information while withholding the actual diagnosis. Each tested LLM serves as a doctor simulator. This setup enables the agent to provide symptom information and respond to inquiries in a manner that mirrors real patient experiences, drawing inspiration from standardized patients in medical education \cite{Barrows1993AnOO}. We conduct systematic evaluation with senior physician to validate the effectiveness of patient agent. The details can be found in Appendix \ref{sec:task2_history}.

%This validated interaction environment ensures robust  for the \benchmark~benchmark, enabling comprehensive evaluation of LLMs' capabilities in conducting medical consultations, formulating relevant questions, and gathering critical information for diagnosis and treatment planning.

\subsection{Benchmark Evaluation}
Given the complexity of medical decision-making, we developed a comprehensive evaluation framework that goes beyond simple binary assessments. The standard answer for each stage comprises multiple key points. The evaluated agent's score increases with the number of standard points addressed, while irrelevant content leads to score reduction. For each task, we employ specific evaluation metrics: Task 1 (Specialty Referral) uses accuracy and Intersection over Union (IoU) to compare predictions with ground truth. Task 2 (History-taking) calculates IoU between predicted and ground truth examination items. Task 3 leverages DocLens \cite{xie2024doclens} for assessing image interpretation quality. Task 4 employs a carefully designed prompt based on National Health Commission of China guidelines to evaluate diagnostic accuracy. Task 5 measures IoU between predicted and ground truth examination items. We elaborate the evaluation metrics in \autoref{sec:append_benchmark_construction}.

\section{\benchmark-Agent Framework}

\subsection{Multi-agent System with Feedback}

The \benchmark-Agent framework introduces a multi-agent system that simulates the complex, interconnected nature of medical decision-making. This system integrates specialized agents that contribute distinct expertise to the diagnostic and treatment process. We sketch the main content here and more details please refer to \autoref{sec:agent_append}.

\subsubsection{Agent Roles}

Our framework comprises three agent types:

\noindent\textbf{General Agents:} Task-specific agents recruited based on domain requirements. For specialty referral tasks, these agents possess expertise in medical specialties and triage protocols, enabling effective patient routing. They engage in collaborative discussions that mirror real-world medical consultations.

\noindent\textbf{Summarizing Agent:} Acts as a central coordinator, consolidating insights from general agents into coherent decisions. This agent synthesizes collective expertise and delivers final recommendations, similar to a senior physician's role in clinical settings.

\noindent\textbf{Feedback Agent:} Maintains decision quality through continuous evaluation and refinement. This agent assesses outputs, provides targeted feedback, and initiates improvements when necessary, creating a self-correcting decision process.

\subsubsection{Decision Making with Feedback}

The decision-making process begins with general agents analyzing patient data and engaging in structured discussions. The summarizing agent then consolidates these assessments into preliminary decisions, which undergo review by the feedback agent. When issues are identified, the feedback agent initiates an iterative refinement cycle, with general agents reconsidering their assessments and the summarizing agent adjusting decisions accordingly. This process continues until reaching consensus or completing a preset number of iterations.

%This multi-layered, iterative approach ensures decisions are products of thorough analysis and diverse perspectives. By incorporating feedback and allowing for refinement, the \benchmark-Agent framework mitigates error propagation across tasks, addressing a common challenge in sequential medical decision-making.

\subsection{MedCase-RAG}

To enhance decision-making capabilities, we developed MedCase-RAG, a specialized Retrieval-Augmented Generation system for medical applications. Unlike traditional medical RAG systems, our approach employs structured case representation using feature vectors that encode 12 key clinical dimensions: Age, Sex, Chief Complaint, Symptoms, Medical History, Physical Examination, Laboratory Tests, Imaging Reports, Diagnosis, Treatment Plan, Primary Department, and Secondary Department. This feature vectors is extracted by text-embedding-ada-002 from openai. When processing new cases, MedCase-RAG performs similarity searches, identifying the three most similar cases using cosine similarity metrics. This approach provides agents with comprehensive reference points for decision-making.
The MedCase-RAG also features dynamic knowledge base expansion, incorporating resolved cases as pseudo-data. This mechanism enables continuous learning from new clinical experiences, adapting to evolving medical knowledge and practices. 

\begin{table*}[t]
	\renewcommand\arraystretch{1.1}
	\centering
	\caption{\textbf{Evaluation of various LLM-based agent in \benchmark.} The best performance for each task is highlighted in \textbf{bold}.}
	\label{tab:performance_result}
	\resizebox{\textwidth}{!}{
		\begin{tabular}{llccccccc}
			\toprule
			\multirow{2}{*}{Framework} & \multirow{2}{*}{Methods} & \multicolumn{2}{c}{Specialty Referral} & \multirow{2}{*}{History-taking} & \multirow{2}{*}{Examination} & \multirow{2}{*}{Diagnosis} & \multirow{2}{*}{Treatment} & \multirow{2}{*}{Average} \\
			\cmidrule(lr){3-4}
			& & Level 1 & Level 2 & & & & & \\ 
			\midrule
			\multirow{11}{*}{\makecell[l]{Base LLM}} 
			& GPT-4o-mini \cite{openai2023gpt4v} & 0.5449 & 0.2871 & 0.3399 & 0.5112 & 0.4177 & 0.3930 & 0.4156 \\
			& GPT-o3 & 0.5495 & 0.2866 & 0.3493 & - & 0.4891 & 0.3343 & 0.4017 \\
			& Claude-3.5-sonnet \cite{team2023gemini} & 0.5681 & 0.3050 & 0.3562 & 0.5018 & 0.4207 & 0.4053 & 0.4262 \\
			& MedGemma \cite{sellergren2025medgemma} & 0.5063 & 0.1975 & 0.3315 & 0.6324 & 0.4668 & 0.3558 & 0.4105 \\
			& LLaVA \cite{liu2024visual} & 0.3240 & 0.0730 & 0.3182 & 0.5165 & 0.3667 & 0.1060 & 0.2841 \\
			& Qwen2 \cite{wang2024qwen2} & 0.4975 & 0.2215 & 0.4226 & 0.4829 & 0.4530 & 0.2193 & 0.3828 \\
			& InternVL2 \cite{wang2024qwen2} & 0.4811 & 0.1935 & 0.4645 & 0.4490 & 0.4367 & 0.2903 & 0.3859 \\
			& Baichuan \cite{dou2025baichuan} & 0.2959 & 0.0532 & 0.3264 & - & 0.4297 & 0.2591 & 0.2728 \\
			& HuaTuoGPT \cite{zhang2023huatuogpt} & 0.0707 & 0.0207 & 0.3465 & - & 0.4163 & 0.1780 & 0.2064 \\
			& MedReason \cite{wu2025medreason} & 0.4602 & 0.1258 & 0.3315 & - & 0.4741 & 0.2993 & 0.3381 \\
			& FineMedLM-o1 \cite{yu2025finemedlm} & 0.1502 & 0.0124 & 0.3969 & - & 0.2772 & 0.1345 & 0.1942 \\
			\midrule
			\multirow{5}{*}{\makecell[l]{Single-agent}} 
			& Zero-shot & 0.4811 & 0.1935 & 0.3450 & 0.4994 & 0.4572 & 0.2817 & 0.3763 \\
			& Few-shot & 0.5584 & 0.2481 & 0.4870 & 0.3758 & 0.4382 & 0.3553 & 0.4105 \\
			& CoT \cite{wei2022chain} & 0.5698 & 0.1484 & 0.3750 & 0.6396 & 0.4382 & 0.3421 & 0.4189 \\
			& RAG \cite{lewis2020retrieval} & 0.5927 & 0.2467 & 0.4732 & 0.6408 & 0.4167 & 0.4524 & 0.4704 \\
			& Self-consistency \cite{wang2022self} & 0.5143 & 0.2319 & 0.4213 & 0.4144 & 0.4198 & 0.3441 & 0.3910 \\ 
			\midrule
			\multirow{5}{*}{Multi-agent}
			& DyLAN \cite{liu2023dynamic} & 0.4415 & 0.1731 & 0.4434 & 0.4980 & 0.3863 & 0.2983 & 0.3734 \\
			& AutoGen \cite{wu2023autogen} & 0.5228 & 0.2347 & 0.4528 & 0.4559 & 0.4250 & 0.3590 & 0.4084 \\ 
			& MedAgent \cite{tang2023medagents} & 0.3830 & 0.2039 & 0.4454 & 0.4456 & 0.4102 & 0.3673 & 0.3759 \\
			& MDAgent \cite{kim2024adaptive} & 0.2398 & 0.1343 & 0.4240 & 0.4983 & 0.3973 & 0.3620 & 0.3426 \\
			&MDAgent + RAG \cite{kim2024adaptive} &0.4754 &	0.2063 &	0.4412 &	0.5346 &	0.4198 &	0.4371 &	0.4190 \\
			\rowcolor{gray!25} & \textbf{MedChain-Agent} & \textbf{0.5873} & \textbf{0.3505} & \textbf{0.5836} & \textbf{0.6566} & \textbf{0.4807} & \textbf{0.4613} & \textbf{0.5200} \\
			\bottomrule
		\end{tabular}
	}
\end{table*}

\section{Experiments}
\subsection{Experimental Setup}
We split the dataset into training, validation, and testing sets with a ratio of 7:1:2. 
Our study evaluates both single-agent and multi-agent systems. For base LLM, we test two closed-source models (gpt-4o-mini \cite{openai2023gpt4v}, and claude-3.5-sonnet \cite{team2023gemini}) and four open-source models (InternVL2-8b \cite{chen2023internvl}, llava-llama-3-8b-v1\_1 \cite{liu2024visual}, HuaTuoGPT \cite{zhang2023huatuogpt}, and Qwen2-7B-Instruct \cite{wang2024qwen2}), with model weights obtained from official Hugging Face repositories. In the single-agent evaluation, we compare with zero-shot manner, few-shot manner, CoT \cite{wei2022chain}, Self-consistency \cite{wang2022self} and RAG \cite{lewis2020retrieval}. 
In the multi-agent evaluation, we compare \benchmark-Agent against DyLAN \cite{liu2023dynamic}, AutoGen \cite{wu2023autogen}, MedAgent \cite{tang2023medagents} and MDAgent \cite{kim2024adaptive}. 
All agent framework are based on InternVL2-8b \cite{chen2023internvl}. The deployment was conducted using the LMDeploy framework \cite{2023lmdeploy}. All tests executed on NVIDIA A100 GPUs featuring 80GB of memory. To enhance output stability and reliability across all experiments, we consistently set the temperature parameter to 0. The experiments were conducted in Chinese.

\begin{table*}[tbp]
	\renewcommand\arraystretch{1.1}
	\centering
	\caption{\textbf{Performance Comparison of LLM-based Agents across Other Diagnosis Benchmarks.} Implementation of agent frameworks based on InternVL2, evaluated on multiple medical benchmarks including MedQA \cite{jin2021disease}, PubMedQA \cite{jin2019pubmedqa}, PathVQA \cite{he2020pathvqa}, and MedBullets \cite{chen2024benchmarking}.}
	\label{tab:performance_result_pre_full}
	\resizebox{0.85\linewidth}{!}{
		\begin{tabular}{llcccccc}
			\toprule
			Framework & Method & MedQA & PubMedQA & PathVQA & MedBullets & Average \\
			\midrule
			\multirow{4}{*}{Single-agent} & Zero-shot & 0.426 & 0.668 & 0.449 & 0.490 & 0.508 \\
			& Few-shot & \textbf{0.477} & 0.648 & 0.448 & \textbf{0.503} & 0.519 \\
			& CoT & 0.470 & 0.714 & 0.465 & 0.500 & 0.537 \\
			& Self-consistency & 0.460 & 0.688 & 0.482 & 0.500 & 0.533 \\
			\midrule
			\multirow{5}{*}{Multi-agent} & MedAgents & 0.501 & 0.622 & 0.569 & 0.435 & 0.532 \\
			& MDAgents & 0.435 & 0.744 & 0.582 & 0.422 & 0.546 \\
			& AutoGen & 0.395 & 0.656 & 0.568 & 0.448 & 0.517 \\
			& DyLAN & 0.414 & 0.610 & 0.540 & 0.448 & 0.503 \\
			& MedChain-Agent & 0.462 & \textbf{0.746} & \textbf{0.621} & 0.474 & \textbf{0.576} \\
			\bottomrule
		\end{tabular}
	}
\end{table*}

\subsection{Benchmark Performance Results}
The results of our evaluation in the \benchmark~ are presented in \autoref{tab:performance_result}. Our analysis yields two significant insights: 

\noindent \textit{(1) Sequential decision-making tasks continue to pose significant challenges, even for advanced models.} For instance, within the single-agent frameworks, GPT-4o-mini and InternVL2 achieve average scores of 0.4156 and 0.3859, respectively. These results indicate that despite their sophistication, these models struggle to maintain consistent performance across the sequential stages of clinical decision-making, highlighting the inherent difficulty of these tasks.

%\noindent \textit{(2) Multi-agent frameworks based on InternVL2, such as DyLAN, MedAgent and MDAgent, exhibit inferior performance compared to their single-agent counterparts.}  This degradation suggests that traditional multi-agent approaches may exacerbate error propagation in sequential decision-making processes, leading to reduced overall performance. AutoGen can improve the average performance. In contrast, our proposed {MedChain-Agent} significantly outperforms these multi-agent methods, achieving an average score of 0.5269. This improvement demonstrates that MedChain-Agent effectively mitigates error propagation, enhancing decision quality and reliability in clinical settings.

\noindent \textit{(2) The integration of the MedChain-Agent framework with open-source LLMs demonstrates significant superiority over proprietary models like GPT-4o-mini.} The substantial performance gain observed with MedChain-Agent (average score of 0.5200) implies that our framework can leverage the strengths of open-source LLMs to achieve superior outcomes. This suggests that open-source models, when enhanced with our framework, are not only competitive but can also excel in handling intricate medical decision-making tasks.

\subsection{Evaluation in Existing Diagnosis Datasets}

Moreover, we evaluate MedChain-Agent on several well-established medical QA datasets, including MedQA \cite{jin2021disease}, PubMedQA \cite{jin2019pubmedqa}, PathVQA \cite{he2020pathvqa}, and MedBullets \cite{chen2024benchmarking}. As shown in \autoref{tab:performance_result_pre_full}, our framework demonstrates strong performance compared to both single-agent baselines and multi-agent alternatives. MedChain-Agent achieves the highest average score (0.576). This consistent superiority across various medical QA benchmarks further validates the effectiveness of our framework, even on simpler, more structured tasks that differ from our real-world clinical scenarios.

\subsection{Ablation Studies and Discussion}

\begin{table*}[t]
		\renewcommand\arraystretch{1.2}
		\centering
		\caption{\textbf{Ablation Study for Key Components for \benchmark-Agent.} This table presents the performance impact of sequentially removing the Feedback mechanism and MedCase-RAG from the full \benchmark-Agent framework.}
		\label{tab:ablation_study}
		\resizebox{\textwidth}{!}{
			\begin{tabular}{ccccccccc}
        \toprule
        \multirow{2}{*}{Feedback} &  \multirow{2}{*}{MedCase-RAG} & \multicolumn{2}{c}{Specialty referral}  & \multirow{2}{*}{History-taking} & \multirow{2}{*}{Examination} & \multirow{2}{*}{Diagnosis} & \multirow{2}{*}{Treatment} & \multirow{2}{*}{Average} \\ \cline{3-4}
        & & Level 1 & Level 2 &  &  &  &  & \\ \hline
        &  & 0.5523 & 0.2228 & 0.3285 & 0.6369 & 0.4724 & 0.3915 & 0.4341 \\
        \checkmark &  & 0.5739 & 0.2906 & 0.4222 & 0.6377 & 0.4299 & 0.4209 & 0.4692 \\
        & \checkmark & \textbf{0.5928} & 0.3353 & 0.5801 & 0.6488 & 0.4699 & 0.4568 & 0.5140 \\
        \checkmark & \checkmark & 0.5873 & \textbf{0.3505} & \textbf{0.5836} & \textbf{0.6566} & \textbf{0.4804} & \textbf{0.4613} & \textbf{0.5200} \\
        \bottomrule
    \end{tabular}
		}
	\end{table*}

\begin{table*}[t]
	\renewcommand\arraystretch{1.1}
	\centering
	\caption{\textbf{Generalizability of \benchmark-Agent with Various Base LLM.} We apply the \benchmark-Agent framework to different base LLMs to validate its generalizability. Performance changes are highlighted with \textcolor{green!50}{light green} for improvements and \textcolor{red!50}{light red} for decreases. Results demonstrate that the \benchmark-Agent framework consistently brings performance gains across different base LLMs.}
	\label{tab:llm_ablation}
	\resizebox{\textwidth}{!}{
		\begin{tabular}{lccccccc}
			\toprule
			\multirow{2}{*}{Base LLM} & \multicolumn{2}{c}{Specialty Referral} & \multirow{2}{*}{History-taking} & \multirow{2}{*}{Examination} & \multirow{2}{*}{Diagnosis} & \multirow{2}{*}{Treatment} & \multirow{2}{*}{Average} \\
			\cmidrule(lr){2-3}
			& Level 1 & Level 2 & & & & & \\ 
			\midrule
			GPT-4o-mini & 0.6065 {\scriptsize \colorbox{gray!20!green!20}{(+0.0616)}} & 0.2302 {\scriptsize \colorbox{gray!20!red!20}{(-0.0569)}} & 0.6205 {\scriptsize \colorbox{gray!20!green!20}{(+0.2806)}} & 0.6279 {\scriptsize \colorbox{gray!20!green!20}{(+0.1167)}} & 0.4975 {\scriptsize \colorbox{gray!20!green!20}{(+0.0798)}} & 0.4329 {\scriptsize \colorbox{gray!20!green!20}{(+0.0399)}} & 0.5026 {\scriptsize \colorbox{gray!20!green!20}{(+0.0870)}} \\
			HuaTuoGPT & 0.2554 {\scriptsize \colorbox{gray!20!green!20}{(+0.1847)}} & 0.0097 {\scriptsize \colorbox{gray!20!red!20}{(-0.0110)}} & 0.3722 {\scriptsize \colorbox{gray!20!green!20}{(+0.0257)}} & - & 0.4159 {\scriptsize \colorbox{gray!20!red!20}{(-0.0004)}} & 0.1450 {\scriptsize \colorbox{gray!20!red!20}{(-0.0330)}} & 0.2396 {\scriptsize \colorbox{gray!20!green!20}{(+0.033)}} \\
			Qwen2 & 0.5818 {\scriptsize \colorbox{gray!20!green!20}{(+0.0843)}} & 0.2781 {\scriptsize \colorbox{gray!20!green!20}{(+0.0566)}} & 0.5962 {\scriptsize \colorbox{gray!20!green!20}{(+0.1736)}} & 0.6534 {\scriptsize \colorbox{gray!20!green!20}{(+0.1705)}} & 0.4628 {\scriptsize \colorbox{gray!20!green!20}{(+0.0098)}} & 0.4315 {\scriptsize \colorbox{gray!20!green!20}{(+0.2122)}} & 0.5006 {\scriptsize \colorbox{gray!20!green!20}{(+0.1178)}} \\
			InternVL2 & 0.5873 {\scriptsize \colorbox{gray!20!green!20}{(+0.1062)}} & 0.3505 {\scriptsize \colorbox{gray!20!green!20}{(+0.1570)}} & 0.5836 {\scriptsize \colorbox{gray!20!green!20}{(+0.1191)}} & 0.6566 {\scriptsize \colorbox{gray!20!green!20}{(+0.2076)}} & 0.4806 {\scriptsize \colorbox{gray!20!green!20}{(+0.0439)}} & 0.4613 {\scriptsize \colorbox{gray!20!green!20}{(+0.1710)}} & 0.5200 {\scriptsize \colorbox{gray!20!green!20}{(+0.1341)}} \\
			\bottomrule
		\end{tabular}
	}
\end{table*}

\begin{table}[t]
	\centering
	\caption{\textbf{The ablation study for three key characteristics in \benchmark.} This table presents the impact of personalization, interactivity, and sequentiality on Diagnosis and Treatment tasks. The arrows $\uparrow \downarrow$ next to settings indicate expected performance change direction. The arrows next to results show actual changes (highlighted in $\colorbox{gray!20}{gray}$ when matching expectations).}
	
	\begin{minipage}{0.48\linewidth}
		\centering
		\resizebox{\linewidth}{!}{
			\begin{tabular}{llcc}
				\toprule
				\midrule
				Setting & Model & Diagnosis & Treatment \\
				\midrule
				\multirow{5}{*}{Full} & MedAgent & 0.4106  & 0.3673 \\
				& MDAgent & 0.3959  & 0.3620  \\
				& gpt-4o-mini & 0.4157 & 0.3930 \\
				& InternVL2 & 0.4378	&0.4472\\
				& \benchmark-Agent & 0.4802   & 0.4613 \\
				\midrule
				\multirow{5}{*}{a) w/o Person. $\downarrow$} & MedAgent & 0.3075 $\colorbox{gray!20}{$\downarrow$}$& 	0.3754 $\uparrow$ \\
				& MDAgent & 0.3283$\colorbox{gray!20}{$\downarrow$}$& 	0.3109 $\colorbox{gray!20}{$\downarrow$}$ \\
				& gpt-4o-mini & 0.3906$\colorbox{gray!20}{$\downarrow$}$	& 0.3406$\colorbox{gray!20}{$\downarrow$}$ \\
				& InternVL2 & 0.3702$\colorbox{gray!20}{$\downarrow$}$	& 0.2527$\colorbox{gray!20}{$\downarrow$}$ \\
				& \benchmark-Agent & 0.4159 $\colorbox{gray!20}{$\downarrow$}$ & 0.4310 $\colorbox{gray!20}{$\downarrow$}$ \\
				\midrule
				\bottomrule
		\end{tabular}}
	\end{minipage}
	\hfill
	\begin{minipage}{0.48\linewidth}
		\centering
		\resizebox{\linewidth}{!}{
			\begin{tabular}{llcc}
				\toprule
				\midrule
				Setting & Model & Diagnosis & Treatment \\
				\midrule
				\multirow{5}{*}{b) w/o Seq. $\uparrow$} & MedAgent & 0.4030 $\downarrow$& 0.4456$\colorbox{gray!20}{$\uparrow$}$  \\
				& MDAgent & 0.4497 $\colorbox{gray!20}{$\uparrow$}$ & 0.4418$\colorbox{gray!20}{$\uparrow$}$  \\
				& gpt-4o-mini & 0.4522$\colorbox{gray!20}{$\uparrow$}$  & 0.4423 $\colorbox{gray!20}{$\uparrow$}$ \\
				& InternVL2 & 0.4481 $\colorbox{gray!20}{$\uparrow$}$&	0.2903$\downarrow$ \\
				& \benchmark-Agent & 0.4807 $\colorbox{gray!20}{$\uparrow$}$ & 0.4743 $\colorbox{gray!20}{$\uparrow$}$ \\
				\midrule
				\multirow{5}{*}{c) w/o Inter. $\uparrow$} & MedAgent & 0.3129 $\downarrow$	& 0.4109 $\colorbox{gray!20}{$\uparrow$}$ \\
				& MDAgent & 0.3998	$\colorbox{gray!20}{$\uparrow$}$& 0.3627 $\colorbox{gray!20}{$\uparrow$}$ \\
				& gpt-4o-mini & 0.4663$\colorbox{gray!20}{$\uparrow$}$	& 0.4003 $\colorbox{gray!20}{$\uparrow$}$ \\
				& InternVL2 & 0.4550 $\colorbox{gray!20}{$\uparrow$}$	& 0.3173 $\downarrow$ \\
				& \benchmark-Agent & 0.4634 $\downarrow$ & 0.5207 $\colorbox{gray!20}{$\uparrow$}$ \\
				\midrule
				\bottomrule
		\end{tabular}}
	\end{minipage}
	
	\label{tab:coherence_comparison}
\end{table}

\noindent \textit{(1) Ablation Study for Key components in \benchmark-Agent:} To assess our framework's components, we conduct ablation studies as shown in \autoref{tab:ablation_study}. Both the Feedback mechanism and MedCase-RAG module demonstrate significant individual contributions to overall performance. The Feedback mechanism alone improves the average score from 0.4341 to 0.4692, with notable gains in History-taking and Level 2 Specialty referral. MedCase-RAG shows stronger individual impact, boosting the average score to 0.5140, with substantial improvements in History-taking and Specialty referral tasks. While MedCase-RAG excels in diagnostic phases, the Feedback mechanism appears more beneficial for Treatment tasks. When combined, these components show synergistic effects, achieving the highest average performance (0.5200) and optimal scores across five of six evaluated tasks, validating their complementary roles in enhancing clinical reasoning capabilities.

\noindent \textit{(2) Ablation Study for Three Key Characteristics in \benchmark}: To validate the effectiveness of personalization, interactivity, and sequentiality within our benchmark, we conduct an ablation study as shown in \autoref{tab:coherence_comparison}. We systematically remove each characteristic and observe its impact on model performance across Diagnosis and Treatment tasks. 'w/o Person' means all detailed patient profiles are omitted from the input, resulting in a lack of personalized information that makes correct diagnosis more challenging for the model. 'w/o Seq' means we use the ground truth from the previous stage as input to the next stage, rather than using the model's previous output, which simplifies the benchmark. 'w/o Inter' means we directly provide all patient examination results as input without requiring the agent to autonomously inquire about the patient's condition, also simplifying the benchmark. Removing patient-specific information (\textit{w/o Person.}) consistently degrades performance across all models in diagnosis tasks (with drops ranging from 2.51\% to 10.31\%), demonstrating that personalized information is crucial for accurate clinical decision-making. Interestingly, when removing the sequential dependency between stages (\textit{w/o Seq.}), most models show improved performance, indicating that sequential decision-making poses greater challenges that better reflect real-world clinical scenarios. Similarly, the improved performance observed after removing interactive information gathering (\textit{w/o Inter.}) confirms the effectiveness of interactivity in our benchmark design. These results collectively suggest that while both sequentiality and interactivity make the benchmark more challenging, they are essential components that better simulate the complexity of real-world clinical decision-making processes.

\noindent \textit{(3) Generalizability for \benchmark-Agent:} To evaluate the generalizability of our framework, we apply \benchmark-Agent to various base LLMs as shown in \autoref{tab:llm_ablation}. The results demonstrate substantial performance improvements across most models and tasks. InternVL2 shows the most significant enhancement with a 13.41\% average improvement, followed by Qwen2 (11.78\%) and GPT-4o-mini (8.70\%). Even with HuaTuoGPT, which has relatively lower baseline performance, our framework still achieves a 3.33\% improvement. Notably, the improvements are particularly pronounced in complex tasks such as History-taking and Examination, where gains of up to 28.06\% are observed. These consistent enhancements across diverse models and tasks strongly validate the framework's robust generalizability and effectiveness in medical decision-making scenarios.

\section{Conclusion}
In this paper, we introduced \benchmark, a novel benchmark for evaluating LLM-based agents in clinical decision-making that authentically reflects real-world medical practice through three essential characteristics: personalization, interactivity, and sequentiality. Our comprehensive dataset encompasses 12,163 diverse clinical cases across 19 medical specialties, including 7,338 medical images with corresponding reports, providing a robust foundation for evaluating AI systems in complex healthcare scenarios. To address the challenges presented by this benchmark, we also introduced MedChain-Agent, a multi-agent framework enhanced by feedback mechanisms and MedCase-RAG, which demonstrates superior performance across various clinical tasks. This work establishes new benchmarks for evaluating medical AI systems and provides practical solutions for enhancing their clinical decision-making capabilities. As AI continues to evolve in healthcare applications, frameworks that can navigate the complexities of real-world clinical scenarios will be increasingly valuable for improving patient care and supporting healthcare professionals.

\iffalse
\section*{Acknowledgement}
We are particularly indebted to the administrators of the iiyi website for their generosity in allowing us to utilize their data for our research purposes. 
We would like to acknowledge the assistance provided by Claude-3.5 in proofreading our manuscript for grammatical accuracy and in facilitating the creation of LaTeX tables.
\fi

\section{Limitations}

This paper has two primary limitations that offer avenues for future research:

1) Data Source Diversity: The MedChain benchmark is constructed from 12,163 electronic health records obtained from the Chinese medical website ``iiYi." Although this dataset is extensive and covers 19 medical specialties and 156 sub-categories, it is derived from a single source. Additionally, there exists a notable imbalance in data distribution across different medical specialties, with some departments having significantly more cases than others. This imbalance may introduce bias in model evaluation and limit performance in underrepresented specialties. In our future work, we will incorporate additional data sources from different regions or healthcare systems to further enrich the dataset and address the specialty imbalance.
	
2) Patient Interaction Simulation: In our interactive environment, the patient responses during the history-taking stage are generated by the Gemma 2 language model. While this approach ensures consistency and control in evaluating the LLM-based agent, the real patient interactions can be more varied and complex. Future work could explore more advanced patient simulators or incorporate real dialogue data to capture a wider range of communication styles and behaviors.

\iffalse
\begin{ack}
Use unnumbered first level headings for the acknowledgments. All acknowledgments
go at the end of the paper before the list of references. Moreover, you are required to declare
funding (financial activities supporting the submitted work) and competing interests (related financial activities outside the submitted work).
More information about this disclosure can be found at: \url{https://neurips.cc/Conferences/2025/PaperInformation/FundingDisclosure}.

Do {\bf not} include this section in the anonymized submission, only in the final paper. You can use the \texttt{ack} environment provided in the style file to automatically hide this section in the anonymized submission.
\end{ack}
\fi

\bibliography{custom}
\bibliographystyle{plain}

\medskip

%%%%%%%%%%%%%%%%%%%%%%%%%%%%%%%%%%%%%%%%%%%%%%%%%%%%%%%%%%%%
\clearpage
\appendix

\noindent { \LARGE \textbf{Appendix for \benchmark}}

\smallskip
\smallskip
\smallskip

\noindent\textbf{Abstract.} 

\autoref{sec:append_benchmark_construction} describes the process of standardizing and organizing the dataset for the \benchmark benchmark.

\autoref{sec:agent_append} provides a detailed explanation of the \benchmark-Agent framework, including its implementation, feedback mechanism, and the novel Retrieval-Augmented Generation (RAG) approach used to enhance decision-making.

\autoref{sec:experiment_append} lists the additional details for experiment.

\autoref{sec:related_work} discuss the difference between MedChain and several similar related works.

\section{Benchmark Construction and Evaluation}
\label{sec:append_benchmark_construction}

\subsection{Dataset Standardization}
We employed a combination of large language models and human verification to label data across different tasks. Our methodology involves task-specific prompt construction and output matching to ensure data quality and diversity while maintaining alignment with real clinical case scenarios.

\begin{figure*}
	\centering
	\includegraphics[width=\textwidth]{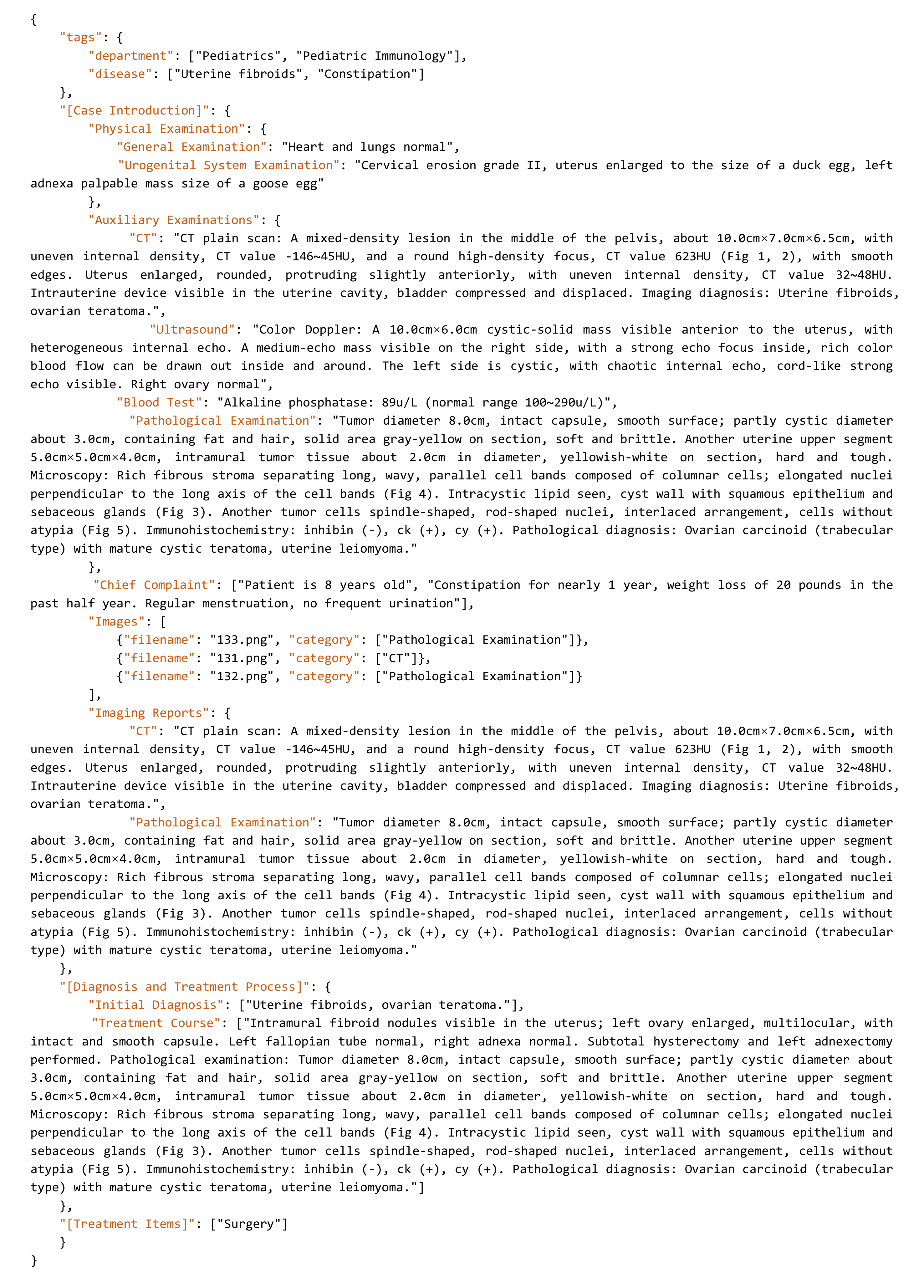}
	\caption{\textbf{Case Report} "77\_Ovarian Carcinoid with Mature Cystic Teratoma: A Case Report."}
	\label{fig:case_show}
\end{figure*}

\begin{figure*}
	\centering
	\includegraphics[width=\textwidth]{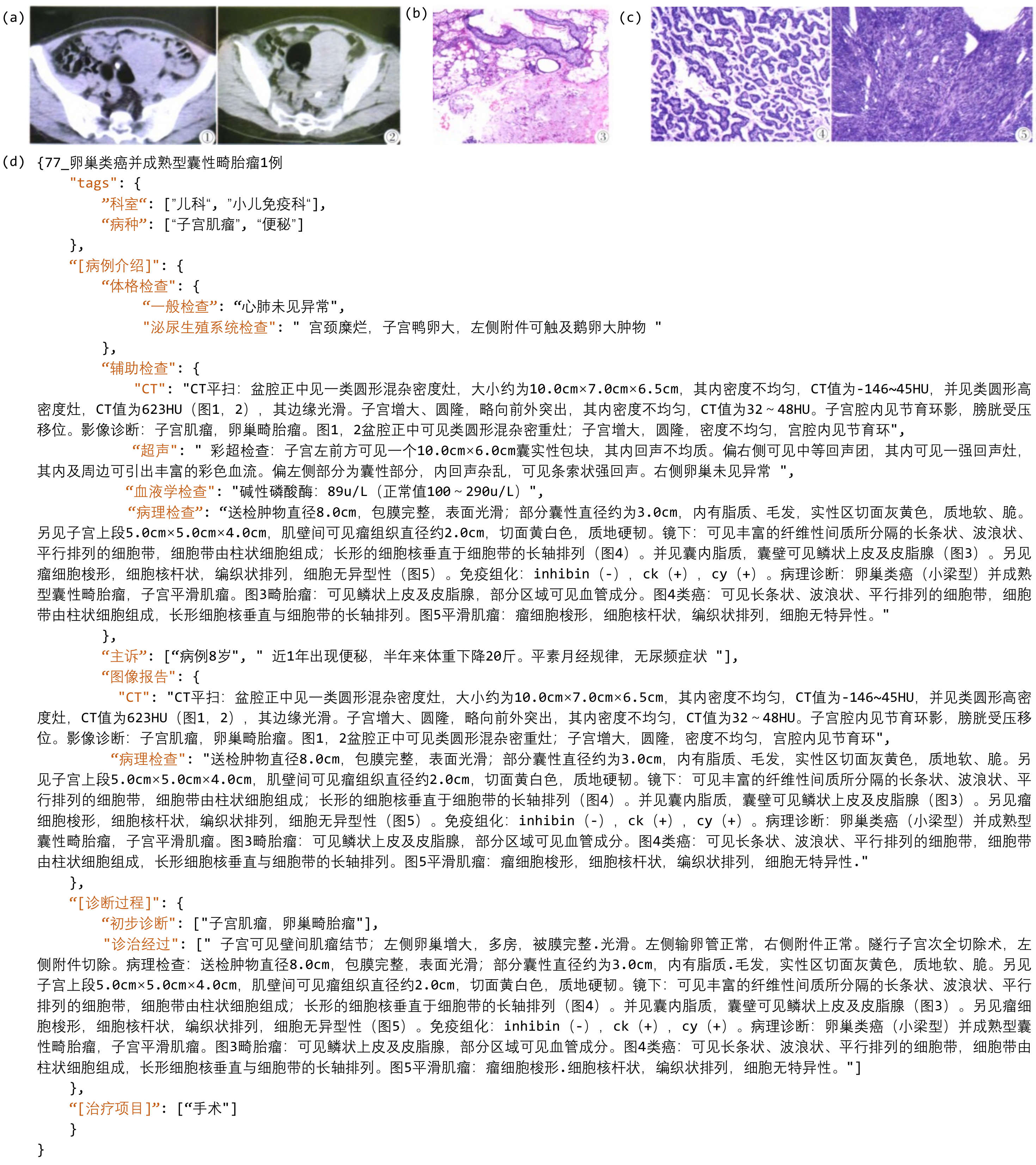}
	\caption{\textbf{Case Report Chinese Version with Corresponding Medical Imaging.} "77\_Ovarian Carcinoid with Mature Cystic Teratoma: A Case Report." (a), (b) and (c) Medical Imaging. (d) Chinese version.}
	\label{fig:case_show_chinese}
\end{figure*}

\begin{figure*}[htbp]
	\centering
	\includegraphics[width=\textwidth]{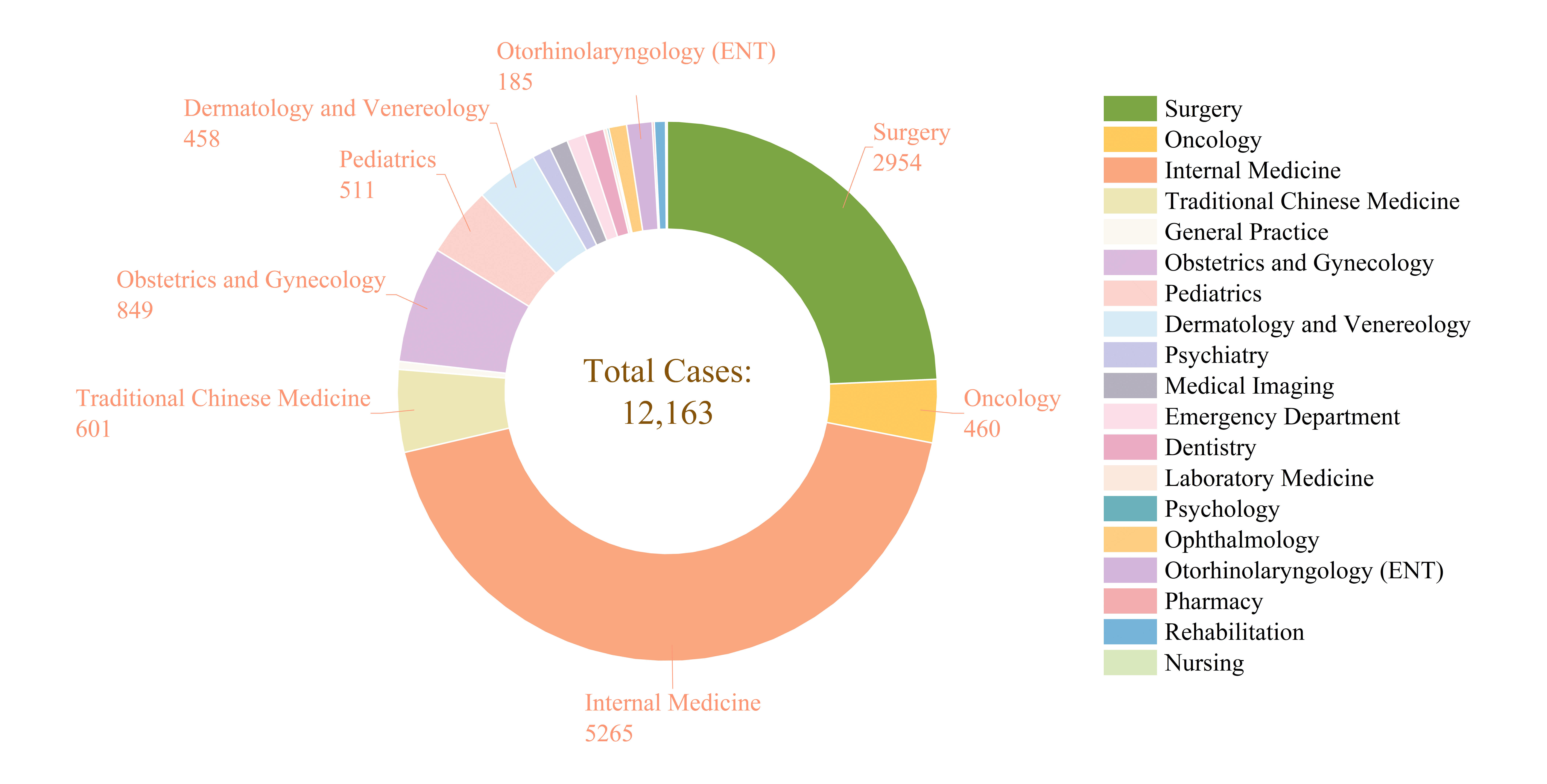}
	\caption{Multi-departmental Distribution}
	\label{fig:department_statistics}
\end{figure*}

To ensure consistency and comparability across the benchmark, we standardized the classification of examination items into two main categories: Physical Examinations and Auxiliary Examinations. Physical Examinations include evaluations of various body systems and general health indicators, while Auxiliary Examinations encompass different imaging techniques and laboratory tests. We utilized GPT-4o to extract and classify examination items from each case, followed by manual verification to ensure accuracy. For medical imaging, we classified images into seven types, and manual review ensured the correctness of the classifications. Additionally, treatment items were extracted and categorized from each case. This standardization process ensures that the dataset is consistent, facilitating accurate and comparable evaluations of LLM performance. \autoref{fig:case_show} and \autoref{fig:case_show_chinese} demonstrate a case after standardization in English and Chinese. \autoref{fig:department_statistics} shows the statistics of different departments.

\subsection{Tasks Details}
This section provides a comprehensive description of the five specific tasks that make up the \benchmark benchmark. Each subsection elaborates on a particular task, detailing its input, output, and evaluation methods. \autoref{fig:benchmark_comparison} visualizes the differences between \benchmark and other methods.

\subsubsection{Task 1: Specialty referral}
The specialty referral task evaluates the LLM's ability to assess the urgency of a patient's condition and determine the appropriate department based on the patient's chief complaint. The input consists of the patient's chief complaint, and the output space includes 19 first-level departments and 156 second-level departments. The LLM must first assign the patient to one of the 19 first-level departments, then to one or more of the 156 second-level departments based on the primary symptoms. 

\textbf{Evaluation metrics} include accuracy for first-level department assignment and Intersection over Union (IoU) for second-level department assignment. IoU is used for second-level departments to account for the possibility of multiple correct assignments and to reflect partial correctness, which can occur in complex cases. This metric better captures the nuanced nature of departmental referrals in clinical practice.

\subsubsection{Task 2: History-taking}
\label{sec:task2_history}
The history-taking  task is designed to simulate doctor-patient communication, where the goal is to obtain relevant information and infer necessary examination items. We employ a multi-agent system to evaluate this process:

\begin{itemize}[itemsep=-5pt, topsep=0pt]
	\item {Doctor Agent}: The LLM being evaluated plays this role, asking questions and suggesting examinations based on the patient's responses.
	\item {Patient Agent}: A local large model (such as Google's gemma2/9b) simulates the patient, responding based on pre-defined case information. This includes the patient's chief complaint, medical history, and examination results.
\end{itemize}

\textbf{Prompt for Patient}: \textit{You are to role-play as a Standardized Patient. Here is your case information: \{patient information\}}.
\textit{You need to answer the doctor's questions directly based on the case information (do not fabricate doctor-patient dialogues). Note that unless the doctor explicitly asks about physical examination and auxiliary examination findings, please do not proactively mention or inquire about physical examination and auxiliary examination related content. If the doctor asks about content that does not exist in the case, please indicate that you don't know and avoid fabricating information. At all times, remember that you are only playing the role of a standardized patient.}

\textbf{Prompt for Doctor}: \textit{You are a doctor. The patient's chief complaint is as follows: \{chief complaints\}}.
\textit{You need to gather more information through conversation with the patient. Physical examination includes: general examination (including height, weight, temperature, blood pressure, pulse, etc.), head, eyes, ears, nose, and throat examination, neck examination (including thyroid, cervical lymph nodes), chest examination (including lungs, heart), abdominal examination, spine and limb examination, skin examination, neurological examination, and genitourinary system examination. Auxiliary examinations include: X-ray, MRI, CT, ultrasound, nuclear medicine imaging, hematological tests, urine tests, stool tests, endoscopic examination, and pathological examination. After obtaining certain information (such as past history and present illness history), please select the physical examinations and auxiliary examinations to inquire about based on the patient's condition. Ask about only one or two items per conversation round. Please inquire about as many examination items as possible (ask at least one item each for physical examination and auxiliary examination) until you can determine the condition. At the end of the conversation, please say "Wishing you a speedy recovery."}

\textbf{Evaluation metrics} is the IoU between the predicted examination items and the ground truth set. This patient-agent design simulates a realistic clinical history-taking process, allowing the doctor-LLM to demonstrate its ability to ask relevant follow-up questions, interpret patient responses, and determine appropriate examinations. The use of a local large model as the Patient Agent ensures consistency in evaluations and improves the reproducibility of the benchmark.

\textbf{Systematic Evaluation}: Furthermore, we conducted a systematic evaluation with a senior physician (10+ years clinical experience) across 10 simulated cases to validate the patient agent's effectiveness. The evaluation focused on three key dimensions: medical history accuracy (match rate between simulated and original EHR data), symptom consistency (temporal coherence of symptom progression), and treatment response fidelity (accuracy of responses to medications and interventions). Each dimension was rated on a 5-point Likert scale (1=poor, 5=excellent). Results demonstrated strong performance across all metrics, with mean scores of 4.0 (variance=1.0) for match rate, 3.9 (variance=1.49) for temporal coherence, and 3.7 (variance=0.81) for treatment response accuracy.

\subsubsection{Task 3: Examination}
The examination task assesses the LLM's ability to analyze medical images and generate corresponding image reports. The input consists of the medical images from each case, and the output is a free-text image report. This task tests the LLM's capability to interpret visual medical data and articulate findings in a clear, professional manner. 

\begin{figure*}
	\centering
	\includegraphics[width=1\linewidth]{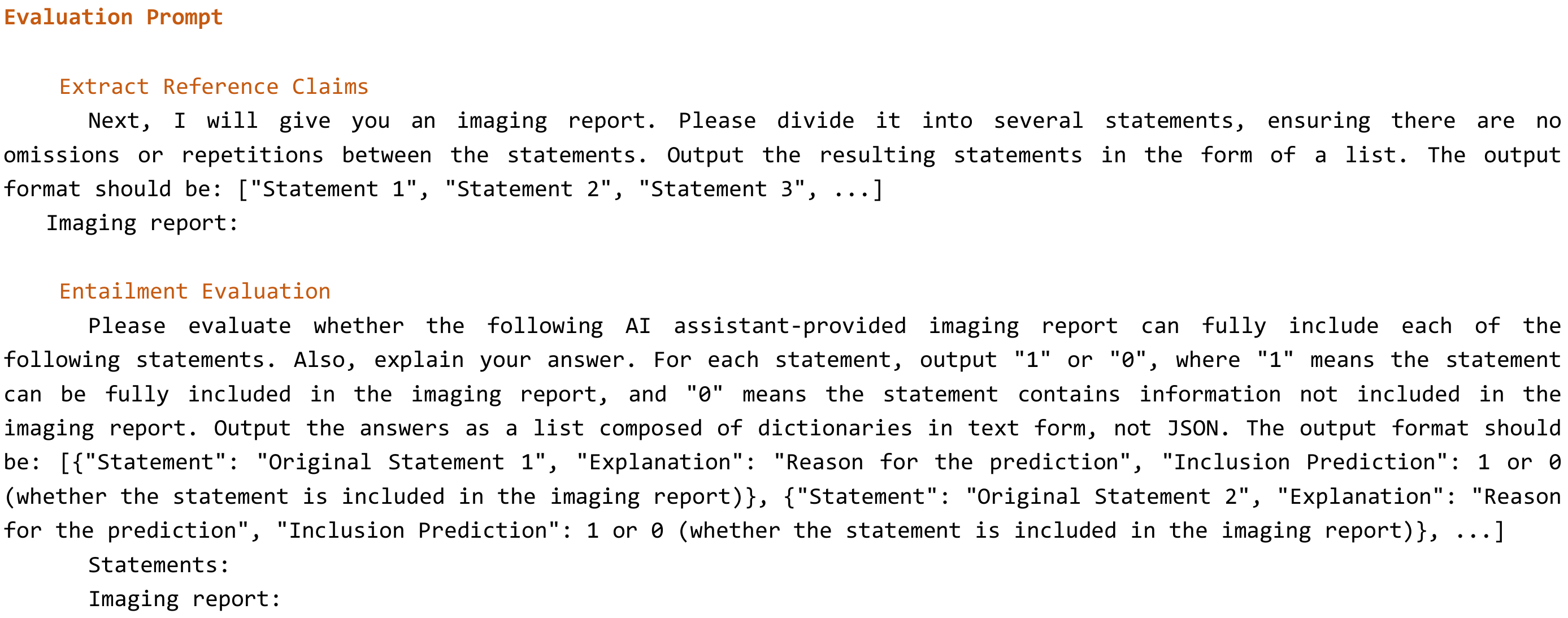}
	\caption{\textbf{The prompt for Evaluation in Task3 Examination.} }
	\label{fig:prompt_evaluation}
\end{figure*}

\textbf{Evaluation metrics:} Given the complexity of medical reports, conventional NLP metrics such as BLEU \cite{papineni2002bleu} and BERTScore \cite{zhang2019bertscore} are inadequate for evaluating this task, as they do not capture whether the generated report aligns with the ground truth in terms of medical claims. Instead, we adopt the Claim Recall metric proposed in DocLens \cite{xie2024doclens} to evaluate the completeness of the generated report. First, GPT-4o-mini is used to extract a list of claims from the original ground truth report. Then, GPT-4o-mini assesses whether the generated report entails these reference claims, and the recall score measures the proportion of claims that are correctly reflected in the generated report. The prompts used for this evaluation can be found in \autoref{fig:prompt_evaluation}.

\subsubsection{Task 4: Diagnosis}
The diagnosis task evaluates the LLM's ability to identify the patient's illness based on the provided symptoms and examination results. The input includes the patient's chief complaint, medical history, examination results, and the department visited. The output is a diagnosis in free-text format. 

\begin{figure*}
	\centering
	\includegraphics[width=1\linewidth]{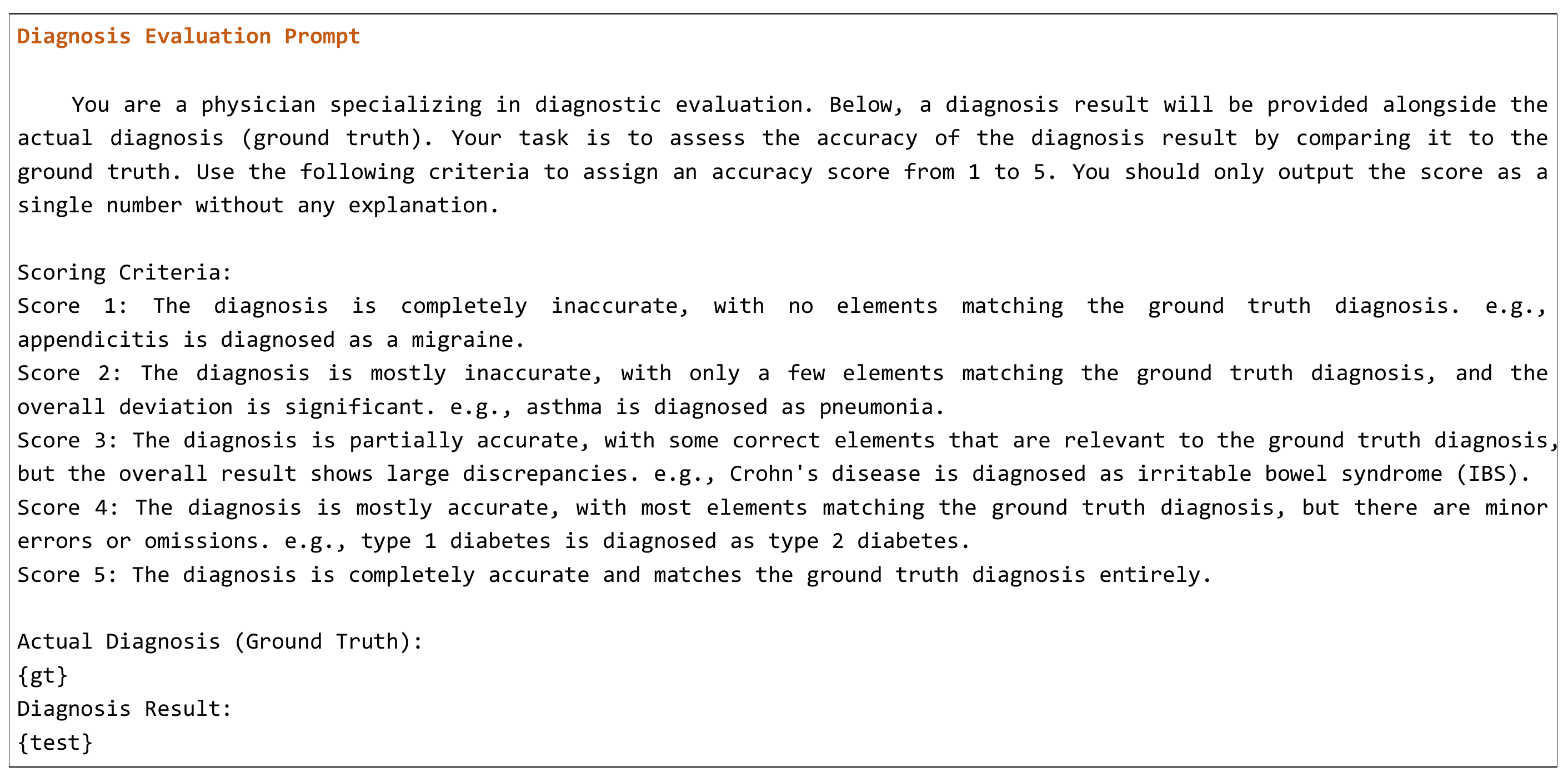}
	\caption{\textbf{The prompt for Evaluation in Task4 Diagnosis.} }
	\label{fig:prompt_diagnosis}
\end{figure*}

\textbf{Evaluation metrics:} We use a five-level grading system to evaluate the accuracy of the diagnosis, ranging from 1 to 5. The prompt we used in LLM-based evaluation is followed the instruction of People's Medical Publishing House \footnote{Diagnostics. 9th Ed. Beijing: People's Medical Publishing House; 2018}. The system accounts for the complex nature of medical diagnoses, where multiple factors may contribute to the patient's condition. The prompt is shown in \autoref{fig:prompt_diagnosis}.

\subsubsection{Task 5: Treatment}
The treatment task involves formulating a treatment plan based on the patient's diagnosis and examination results. The input includes the patient's chief complaint, medical history, examination results, and diagnosis. The output is a set of proposed treatment items. This task tests the LLM's ability to synthesize the gathered information and apply medical knowledge to develop an appropriate treatment plan, considering factors such as drug interactions, treatment protocols, and patient-specific considerations.

\textbf{Evaluation metrics} is the IoU between the proposed treatments and the ground truth treatment set. Since multiple treatment options may be appropriate, IoU allows for partial credit when the LLM suggests a subset of the recommended treatments or proposes additional reasonable treatments that are not part of the ground truth. 

\begin{figure*}
	\centering
	\includegraphics[width=\linewidth]{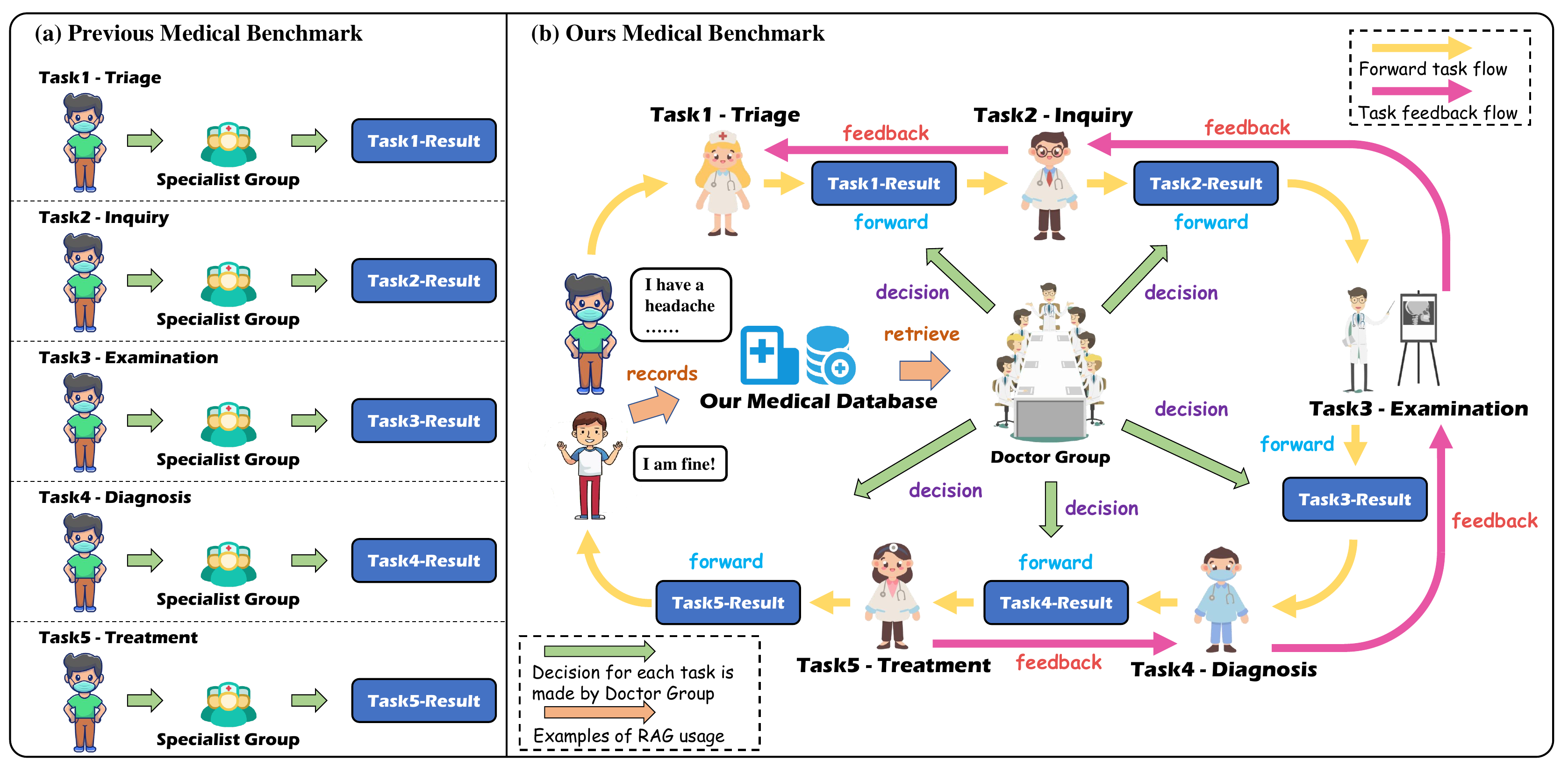}
	\caption{\textbf{Comparison between our benchmark method and previous benchmarks.} In Figure (a), the benchmark methods used in previous work are shown, which are based on specific domain problems and cannot achieve a complete end-to-end medical task workflow. In Figure (b), our proposed benchmark is presented, where we effectively accomplish the full workflow of medical tasks through our proposed MedCase-RAG method and the ACFM mechanism. }
	\label{fig:benchmark_comparison}
\end{figure*}

\subsection{Examination Items in Task 2}
\noindent \textbf{Physical Examination:} General examination (including height, weight, temperature, blood pressure, pulse, etc.), head, eyes, ears, nose and throat examination, neck examination (including thyroid, cervical lymph nodes), chest examination (including lungs, heart), abdominal examination, spine and limb examination, skin examination, neurological examination, urogenital system examination.

\noindent \textbf{Auxiliary Examinations:} X-ray, MRI, CT, ultrasound, nuclear medicine imaging, blood tests, urine tests, stool tests, endoscopy, pathological examination.

\subsection{Treatment Items in Task 5}
Surgery, interventional therapy, medication, chemotherapy, antibiotic therapy, radiation therapy, physical therapy, immunotherapy, psychological therapy, traditional Chinese medicine, gene therapy.

\begin{figure*}
	\centering
	\includegraphics[width=1\linewidth]{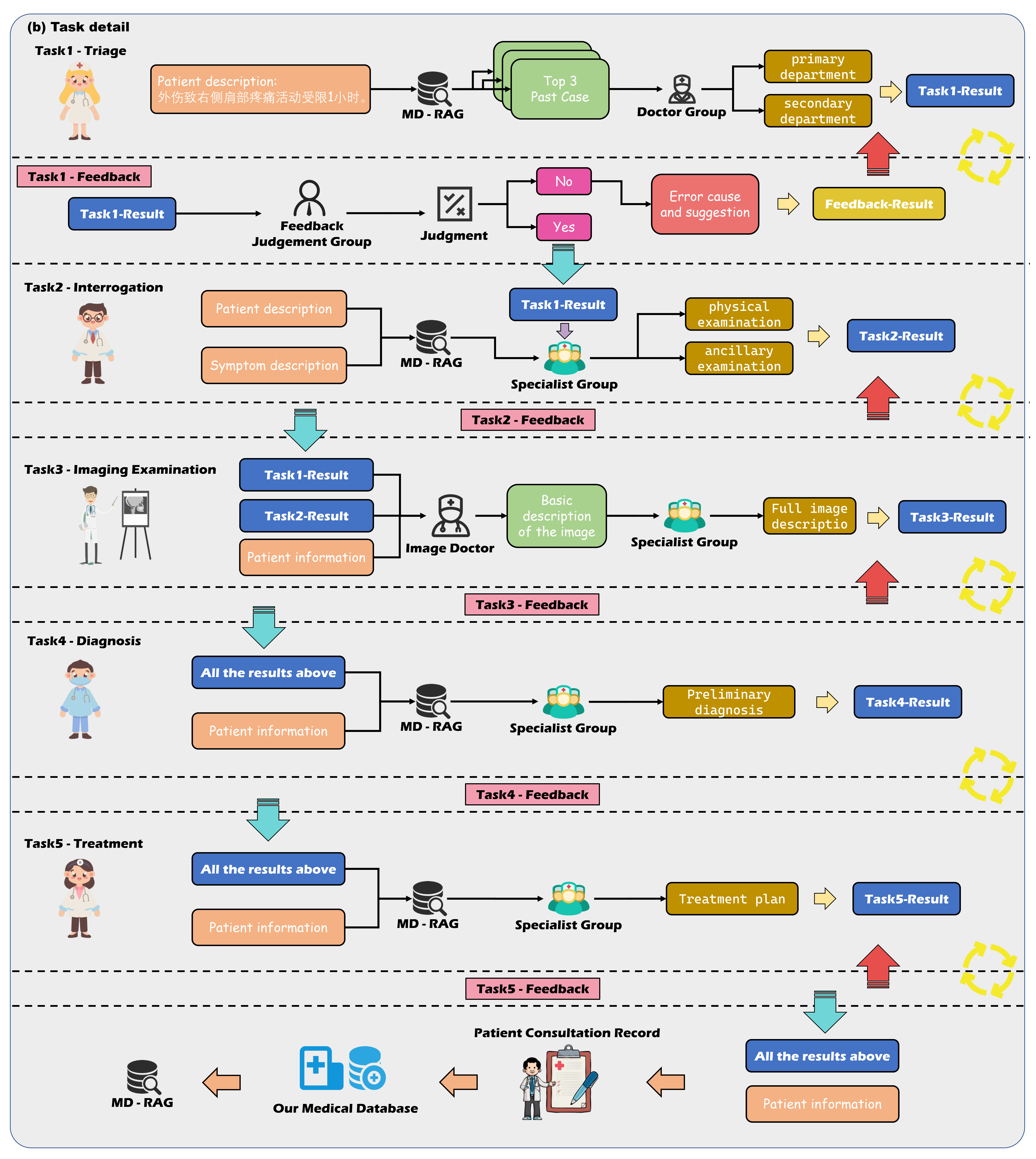}
	\caption{\textbf{The specific agent design for each task in our MedChain-Agent and the ACFM of the entire process are presented.}}
	\label{fig:agent_implementation}
\end{figure*}

\section{\benchmark-Agent}
\label{sec:agent_append}
\subsection{Implementation Details}
The overall process and output flow of our framework are illustrated in \autoref{fig:agent_implementation}. The patient first describes their condition, after which Task 1 performs specialty referral, assigning the patient to the appropriate primary and secondary departments. Once Task 1 is completed, the doctor group consists of specialists from the corresponding primary department, enabling better analytical decision-making. Task 2 involves these specialists prescribing suitable medical examinations based on the patient's current situation. Task 3 focuses on analyzing the patient's imaging data to generate a detailed and comprehensive imaging report. Task 4 consists of the specialists conducting a comprehensive analysis based on the results from Task 2 and the imaging report from Task 3 to provide an initial diagnosis for the patient. Finally, Task 5 involves the specialists formulating an appropriate treatment plan based on all the previous tasks' results and analyses.

Additionally, in the aforementioned description, the decision-making and analytical processes for all tasks are conducted through discussions within the Doctor Group. Specifically, three specialists from the same department first engage in discussions and analysis, after which they summarize their findings and results to the final medical generalist (primary care physician) for the final decision. At each step, relevant similar cases are retrieved from our medical database to assist in decision-making. For Task 1, we will divide the tasks of identifying primary and secondary departments into two subtasks based on the Chain of Thought (CoT) approach: first identifying the primary department, followed by identifying the secondary department. Similarly, for Task 2, we first prescribe specific tests before additional examinations, optimizing the decision-making process.

Once the patient completes the entire medical process, a comprehensive treatment follow-up record will be compiled, including treatment outcomes and reports from each stage, along with the patient's information. This data will then be reintroduced into our medical database. Since these data have not been fully validated (we do not know how reasonable the decisions made for the current cases are), we will treat this data as pseudo-data based on the principles of semi-supervised learning and assign it a lower priority. This means that we will first retrieve data from completely validated sources, and if retrieval fails or data is insufficient, we will resort to retrieving from the pseudo-data.

\subsection{Feedback Mechanism Details}

To date, no work has proposed a multi-agent framework for simulating the entire medical process. Although existing frameworks demonstrate efficiency in specific medical tasks, their performance is suboptimal when linking various medical stages together. In sequential task scenarios, simply concatenating individual tasks is not feasible. Therefore, our full-process task can be regarded as a multi-sequence task, which necessitates a deeper exploration of the issue of error propagation. When a problem arises in the first task, subsequent tasks will analyze and make decisions based on erroneous results, potentially leading to severe impacts on the entire process. 

To address this issue, we have introduced a feedback mechanism within the full-process framework, tightly connecting the current task with all subsequent tasks. After the current task is completed, the output results are evaluated by the physicians of the subsequent tasks (Feedback Judgement Group). Only when the results are confirmed to be accurate will the process advance to the next task; if issues are identified, the reasons for the errors and improvement suggestions will be output and fed back to the current task for re-discussion and decision-making. This process will continue iteratively until consensus is reached on the results or the maximum number of discussion rounds is achieved. 

By implementing a feedback mechanism between each task, we can effectively manage the propagation of errors while significantly enhancing collaboration and communication among agent groups. This mechanism encourages agents to share information in real-time, improving mutual understanding and allowing for rapid strategic adjustments when issues are identified, thereby enhancing the overall flexibility and adaptability of the system. Each agent can better respond to changes in a dynamic environment, facilitating more efficient medical services. This feedback-driven collaborative model lays the foundation for the efficient operation of multi-agent systems, aiding in the provision of precise and reliable medical decision support in complex clinical environments. 

\subsection{RAG Details}

We performed data restructuring on the original dataset, where each case is mapped into feature vectors from 12 dimensions. Among these, "Symptom Description" is identified as the most representative feature of the current patient and is processed through a Text Embedding model for quantification, which is stored in the database for subsequent dense retrieval tasks.

Previous Medical-RAG methods relied on medical question-answering (QA) databases and predominantly used chunked indexing for retrieval. Our approach differs in several ways. Firstly, the content of our foundational medical database is distinct; while most methods have built their databases using medical QA data, ours utilizes a tree structure. We initially categorize patient information according to primary medical departments (in the experimental section, we categorized into 19 primary departments) and then extract and map patient information into a two-dimensional feature representation. This storage design allows for better retention of the patient’s crucial clinical information to assist in decision-making.
Secondly, our retrieval method also diverges from theirs. In our retrieval process, we simulate the everyday practice of physicians. When faced with challenging cases that require decision-making, doctors often refer to past cases as references to enhance current decisions and judgments. We treat these two-dimensional features as the minimal unit of a case, using “Symptom description” as the basis for retrieval. When a new patient arrives, we extract and refine their “Symptom description” feature (approximately 70 characters). We employ direct quantitative matching without tokenizing this feature, using a Text Embedding model for quantification. The resulting feature vector is then compared to each case in our database corresponding to the relevant department through cosine similarity calculations, selecting the top three cases with the highest symptom similarity as the current retrieval results. These results, combined with the current case features, are forwarded to subsequent agents for comprehensive decision-making and judgment, enhancing the agents' output. Since our retrieval method does not utilize document chunking but rather employs direct matching, it retains more matching information, leading to improved matching accuracy.

\subsection{Task Prompts}
\begin{figure*}
	\centering
	\includegraphics[width=1\linewidth]{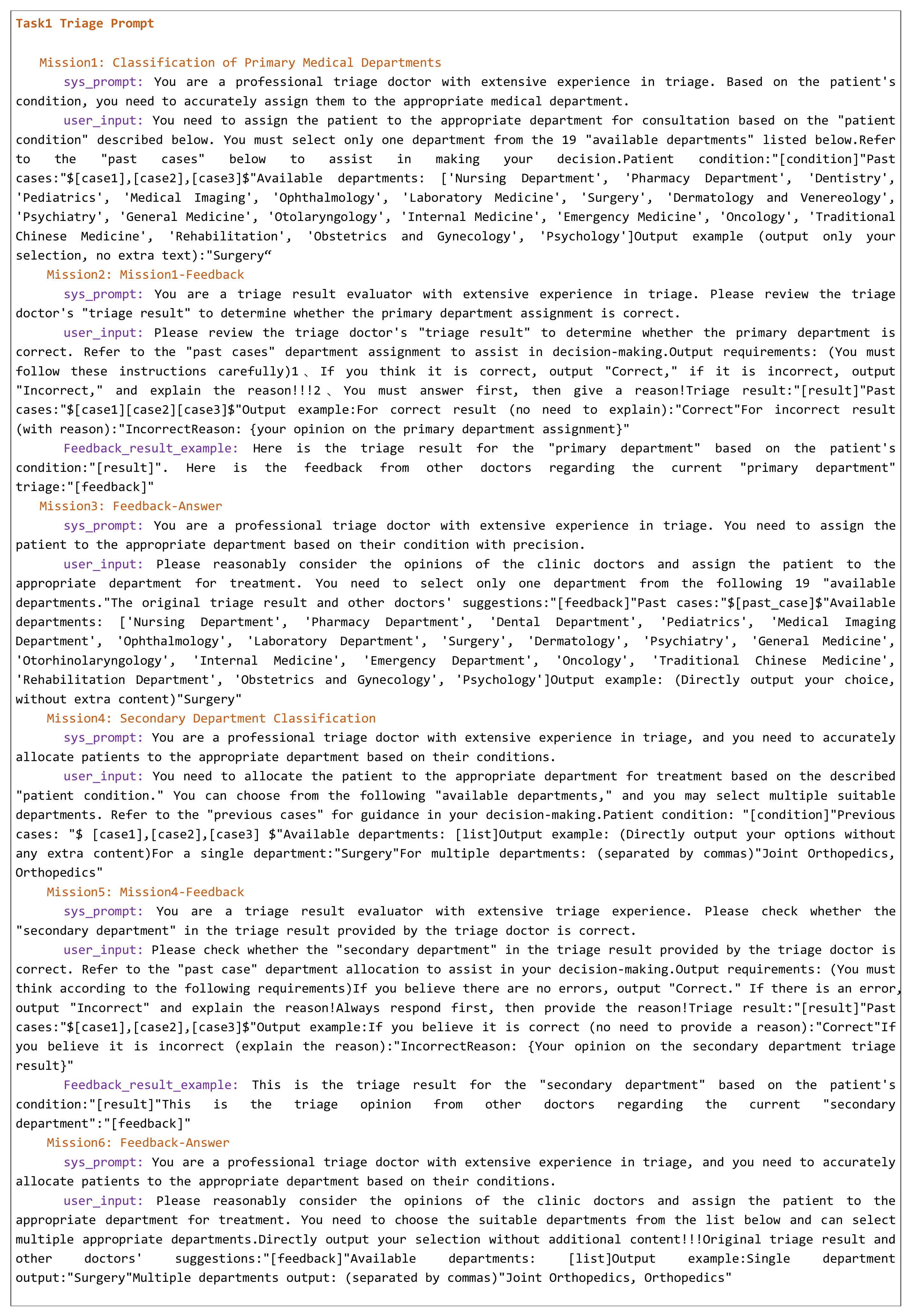}
	\caption{\textbf{The prompt for Task 1 in the MedChain-Agent .} }
	\label{fig:prompt_task1}
\end{figure*}

\autoref{fig:prompt_task1} $\sim$ \autopageref{fig:prompt_task45} showcase prompts for each task and stage within our MedChain-Agent framework.

\begin{figure*}
	\centering
	\includegraphics[width=1\linewidth]{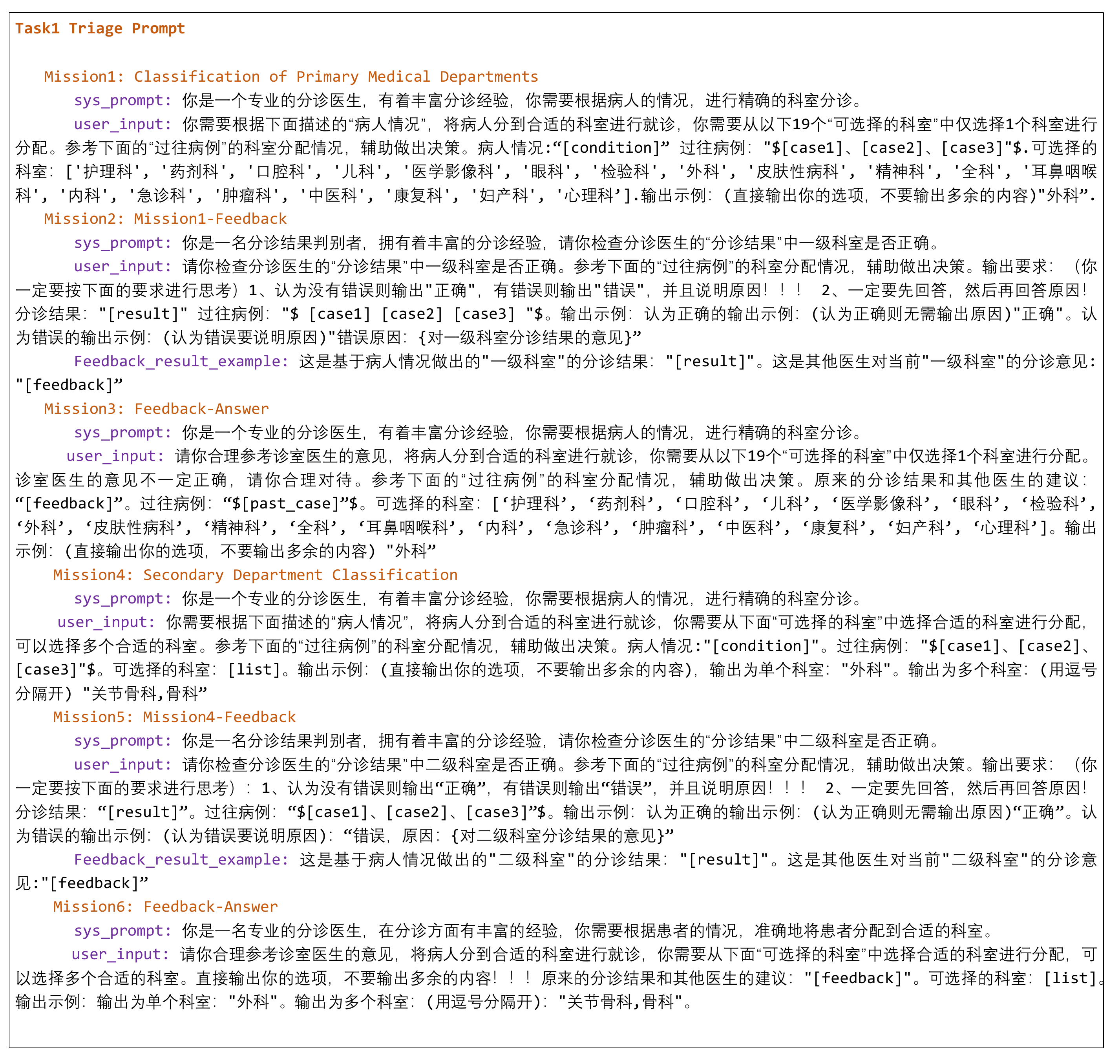}
	\caption{\textbf{The prompt for Task 1 in the MedChain-Agent in Chinese.} }
	\label{fig:prompt_task1_chinese}
\end{figure*}

\begin{figure*}
	\centering
	\includegraphics[width=1\linewidth]{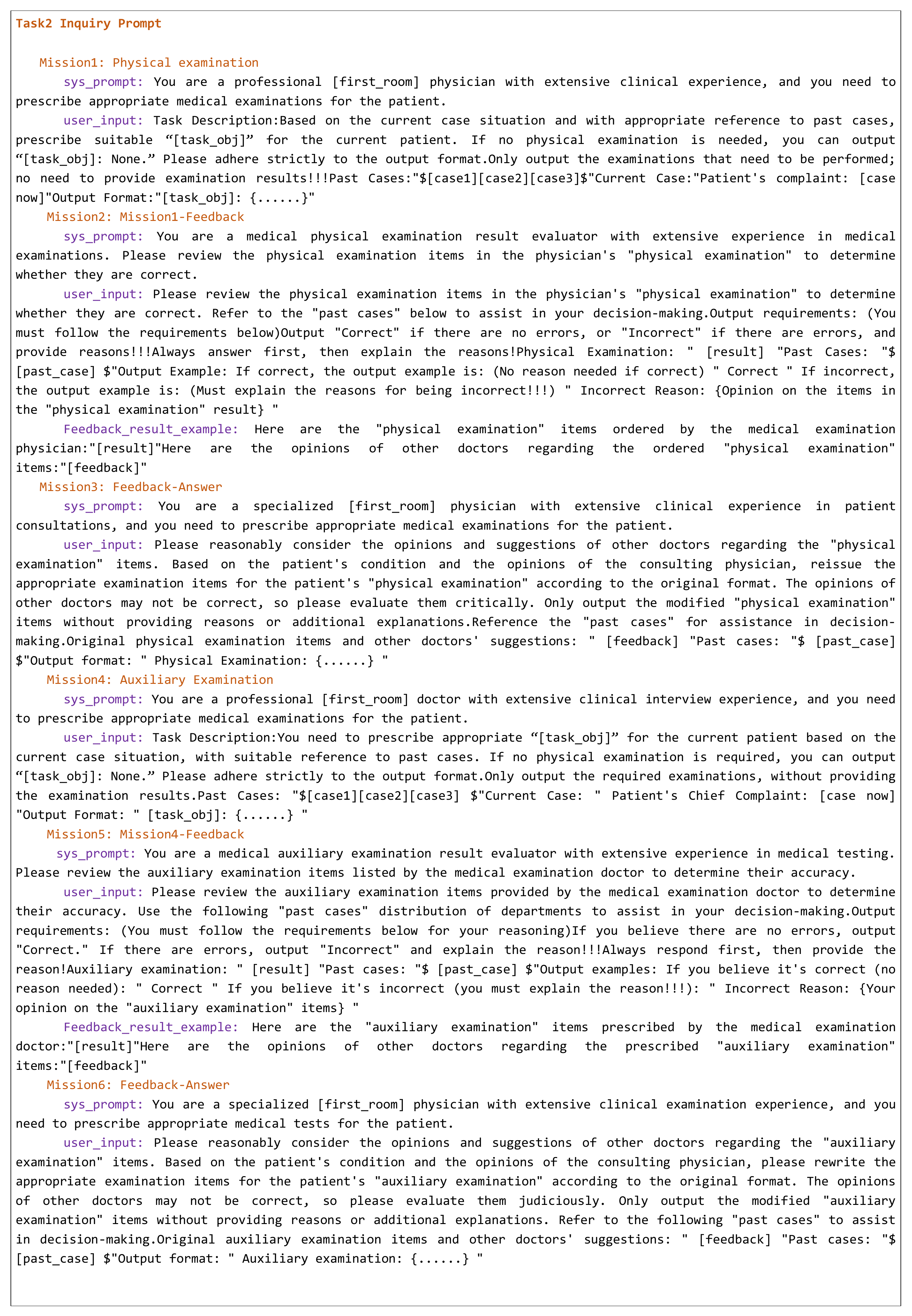}
	\caption{\textbf{The prompt for Task 2 in the MedChain-Agent .} }
	\label{fig:prompt_task2}
\end{figure*}

\begin{figure*}
	\centering
	\includegraphics[width=1\linewidth]{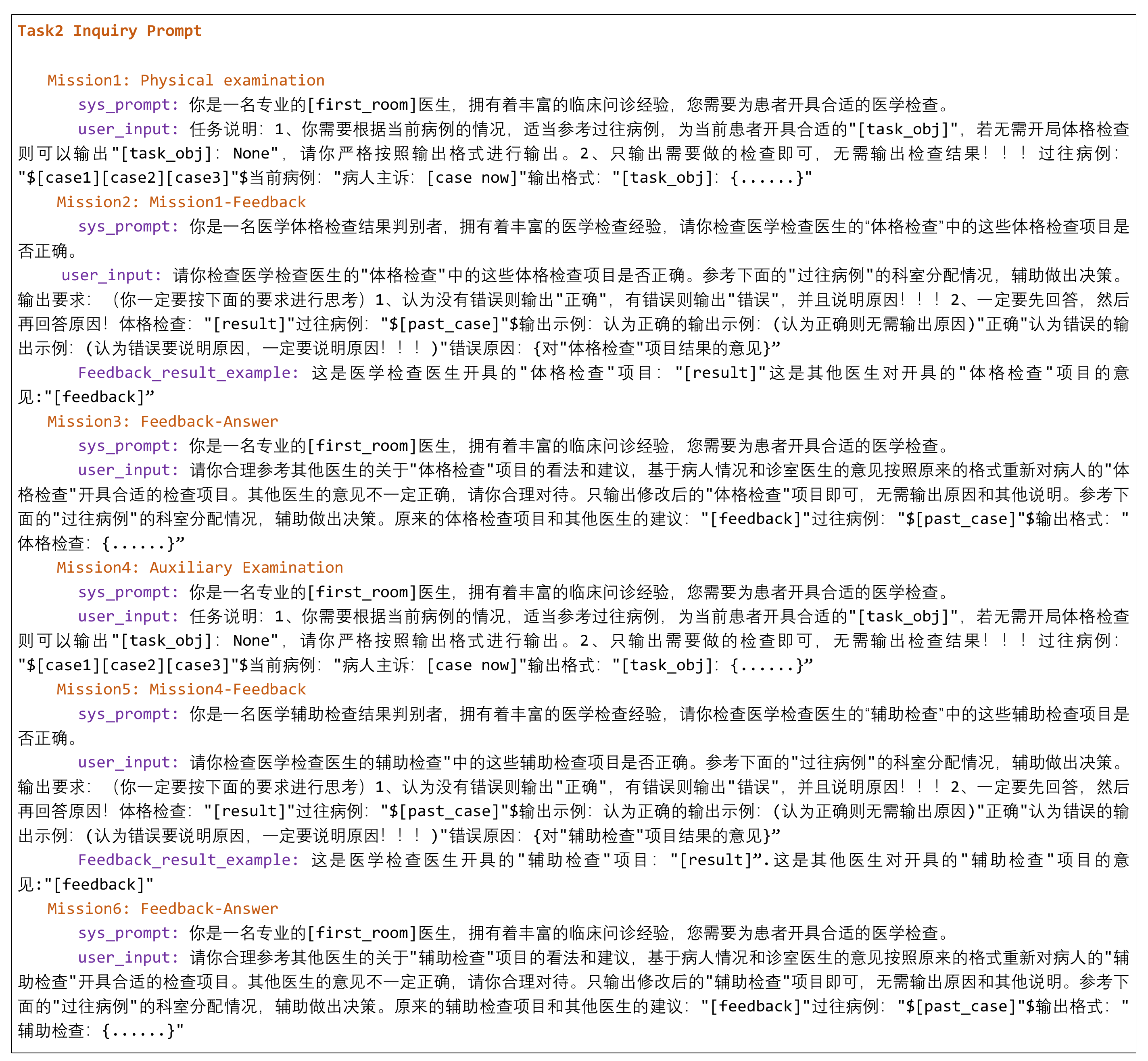}
	\caption{\textbf{The prompt for Task 2 in the MedChain-Agent in Chinese.} }
	\label{fig:prompt_task2_chinese}
\end{figure*}

\begin{figure*}
	\centering
	\includegraphics[width=1\linewidth]{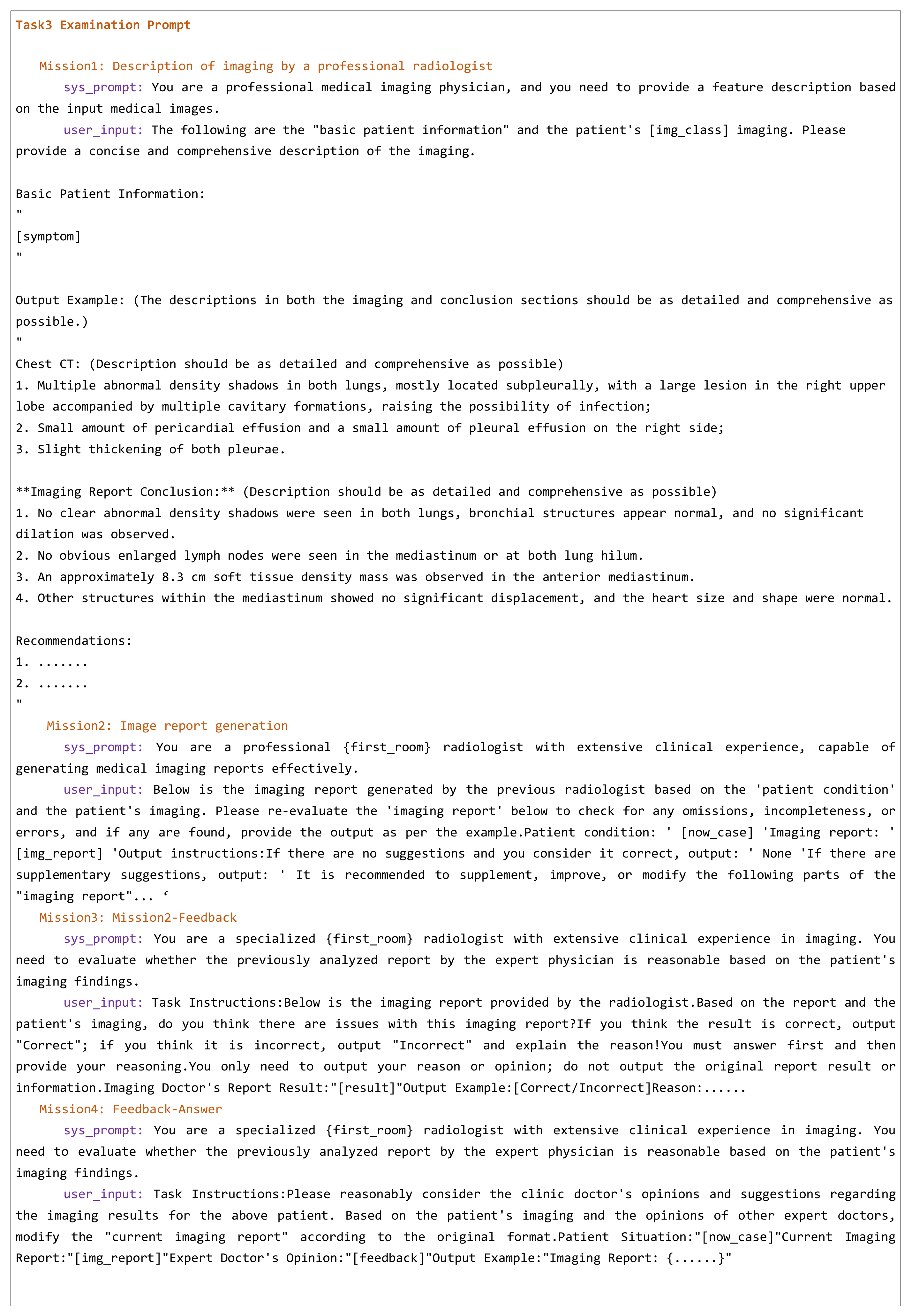}
	\caption{\textbf{The prompt for Task 3 in the MedChain-Agent .} }
	\label{fig:prompt_task3}
\end{figure*}

\begin{figure*}
	\centering
	\includegraphics[width=1\linewidth]{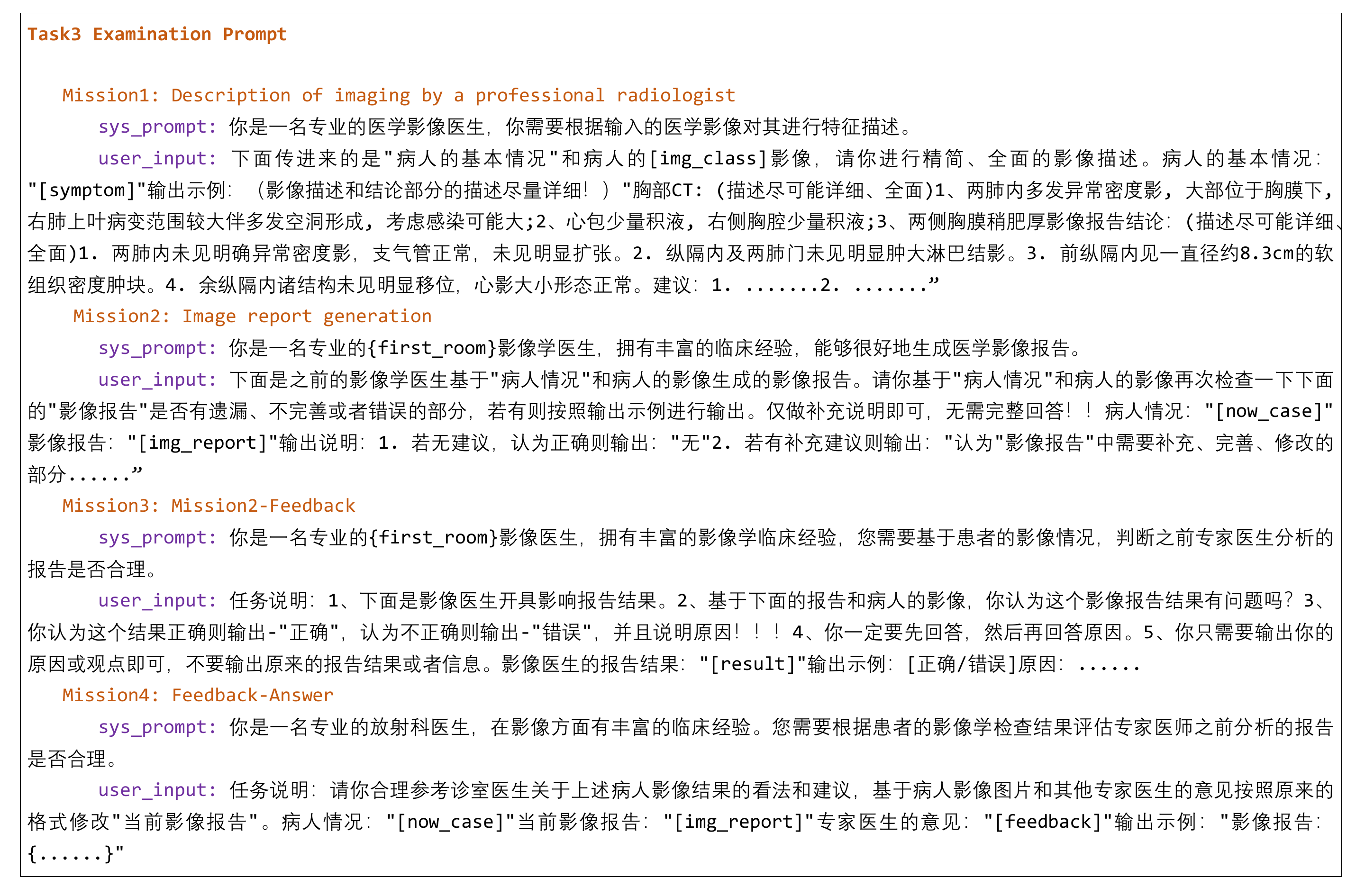}
	\caption{\textbf{The prompt for Task 3 in the MedChain-Agent in Chinese.} }
	\label{fig:prompt_task3_chinese}
\end{figure*}

\begin{figure*}
	\centering
	\includegraphics[width=1\linewidth]{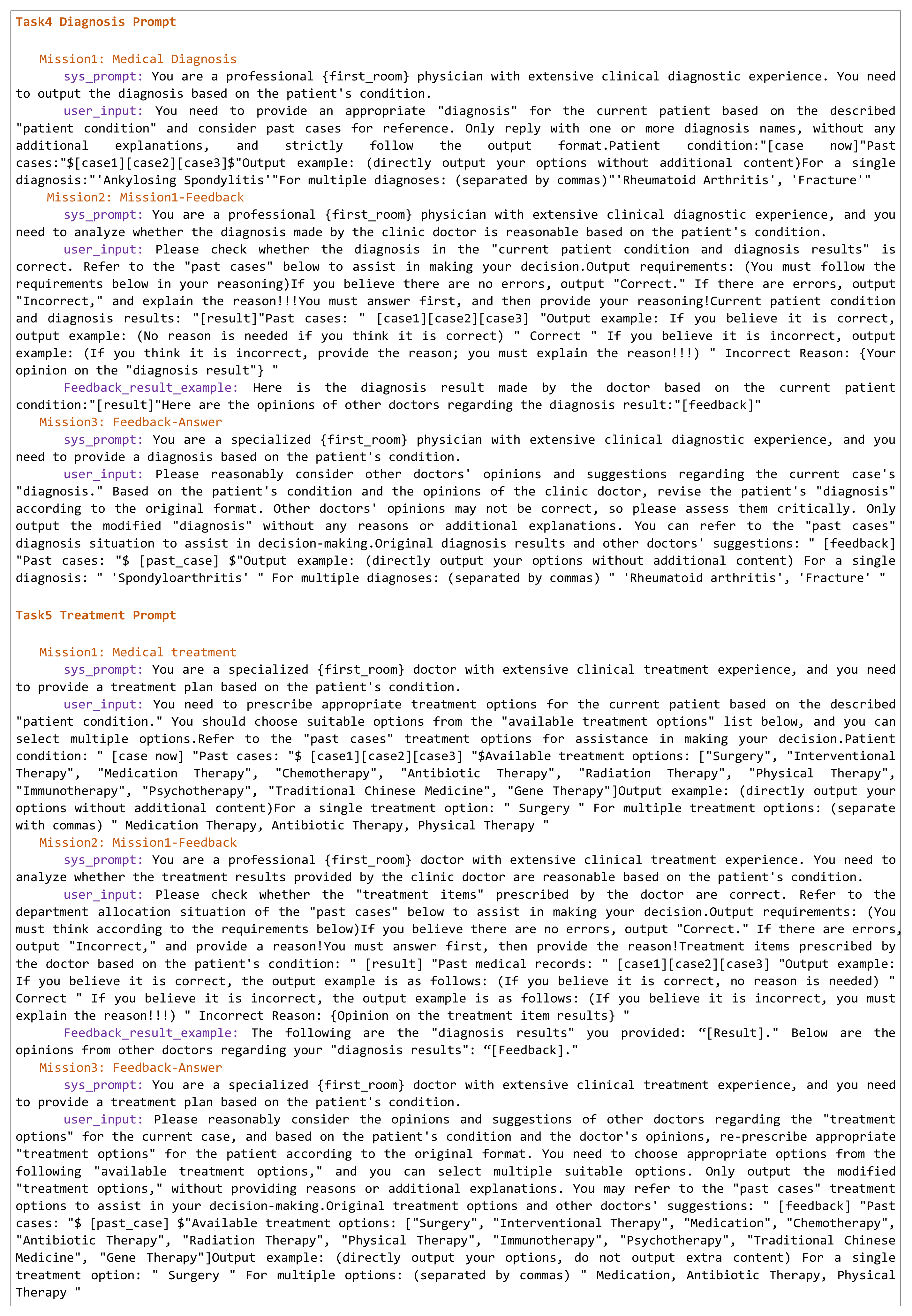}
	\caption{\textbf{The prompt for Task 4 and Task5  in the MedChain-Agent .} }
	\label{fig:prompt_task45}
\end{figure*}

\begin{figure*}
	\centering
	\includegraphics[width=1\linewidth]{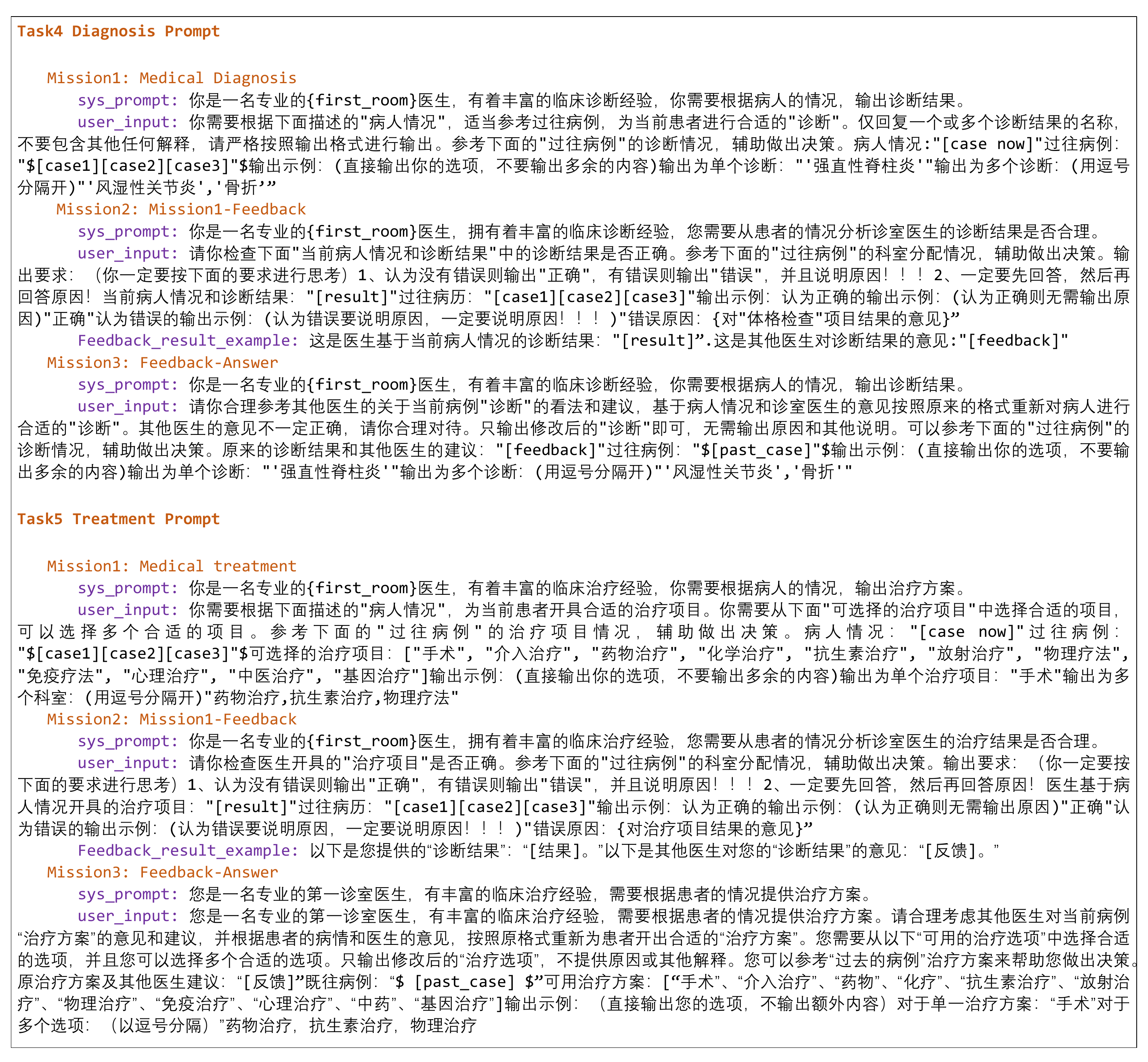}
	\caption{\textbf{The prompt for Task 4 and Task5  in the MedChain-Agent in Chinese.} }
	\label{fig:prompt_task45_chinese}
\end{figure*}

\begin{figure*}
	\centering
	\includegraphics[width=1\linewidth]{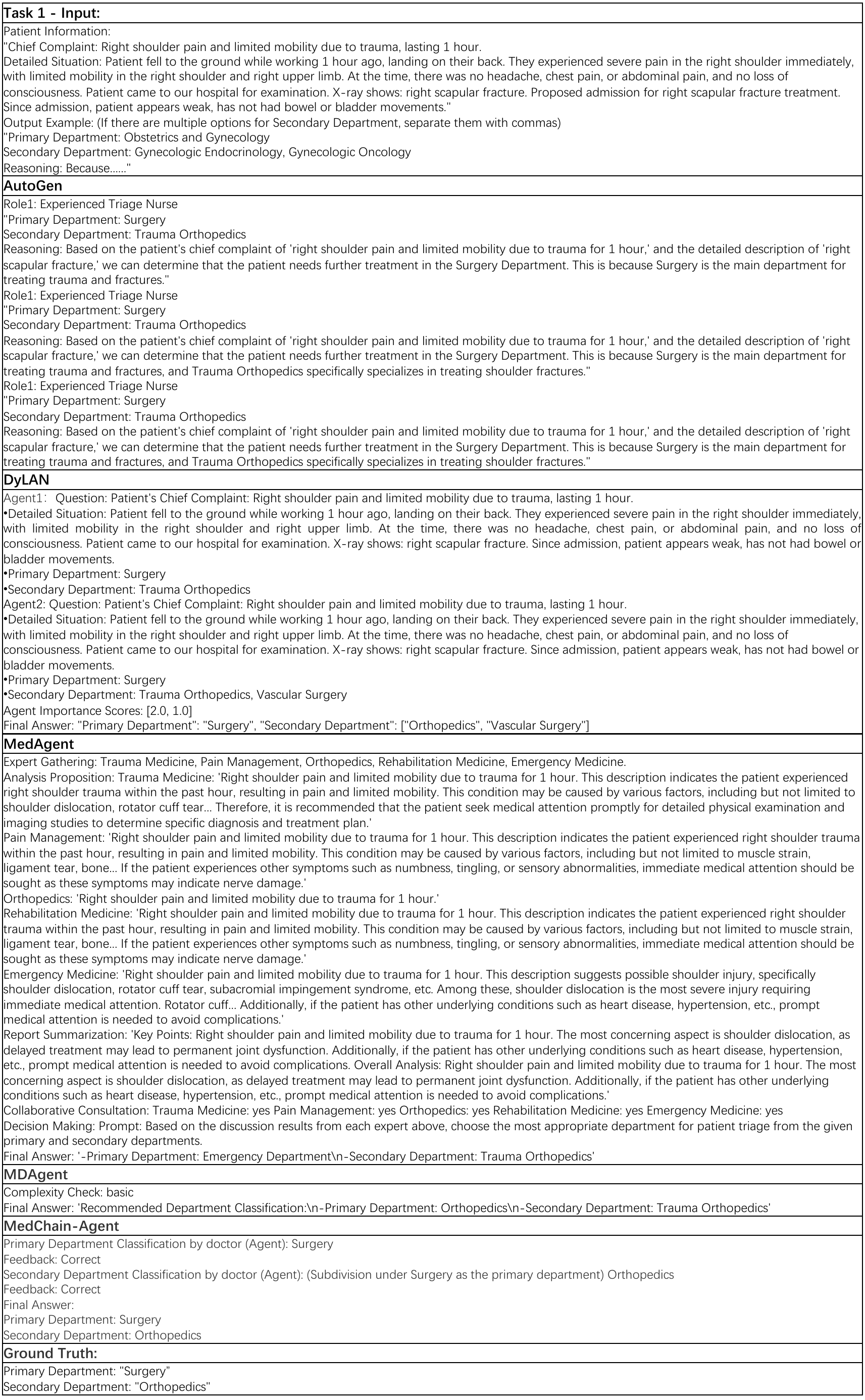}
	\caption{\textbf{The case study among different multi-agent for task1.} }
	\label{fig:case_study_1}
\end{figure*}

\begin{figure*}
	\centering
	\includegraphics[width=1\linewidth]{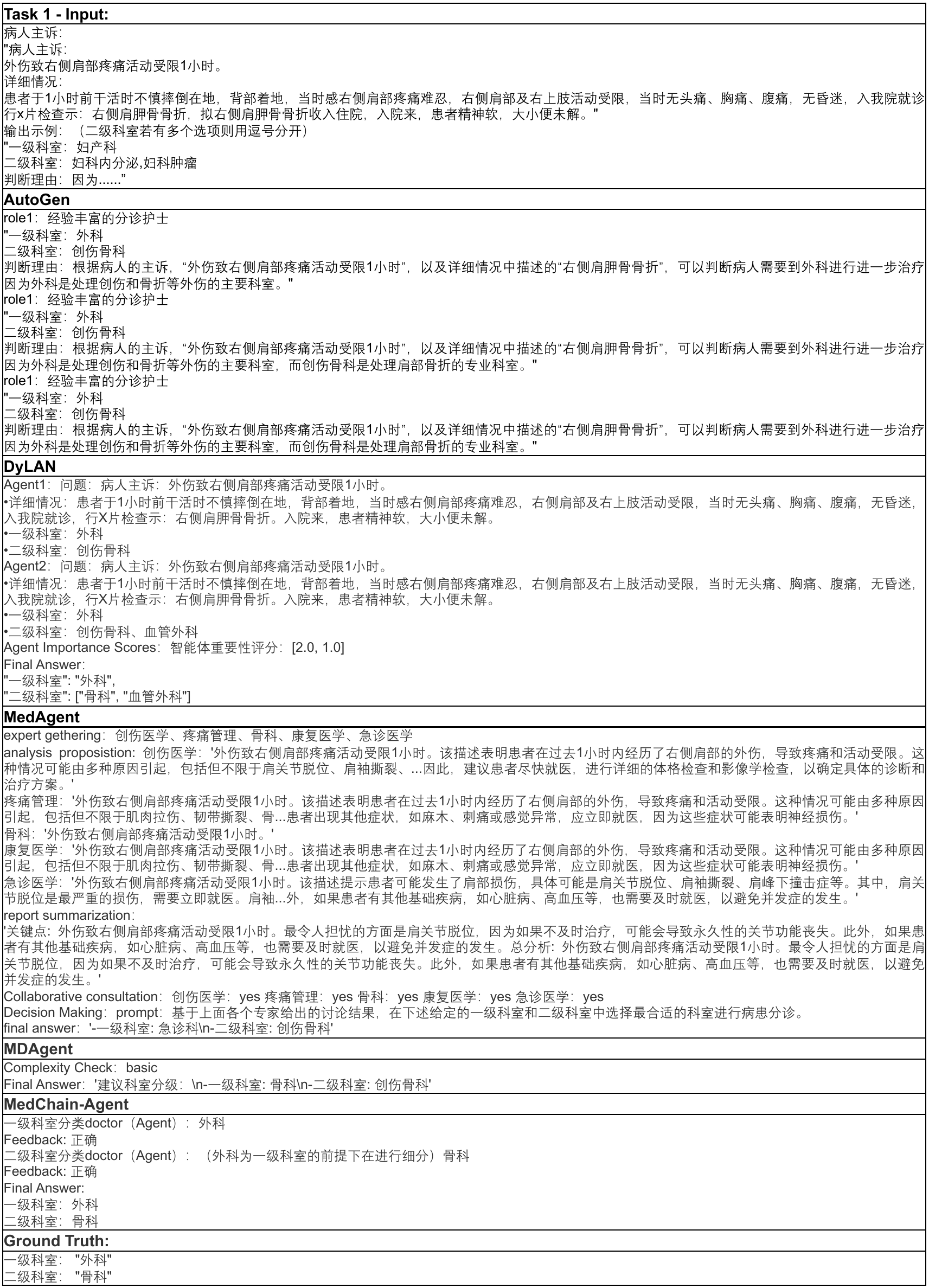}
	\caption{\textbf{The case study among different multi-agent for task 1in Chinese.} }
	\label{fig:case_study_1_chinese}
\end{figure*}

\begin{figure*}
	\centering
	\includegraphics[width=1\linewidth]{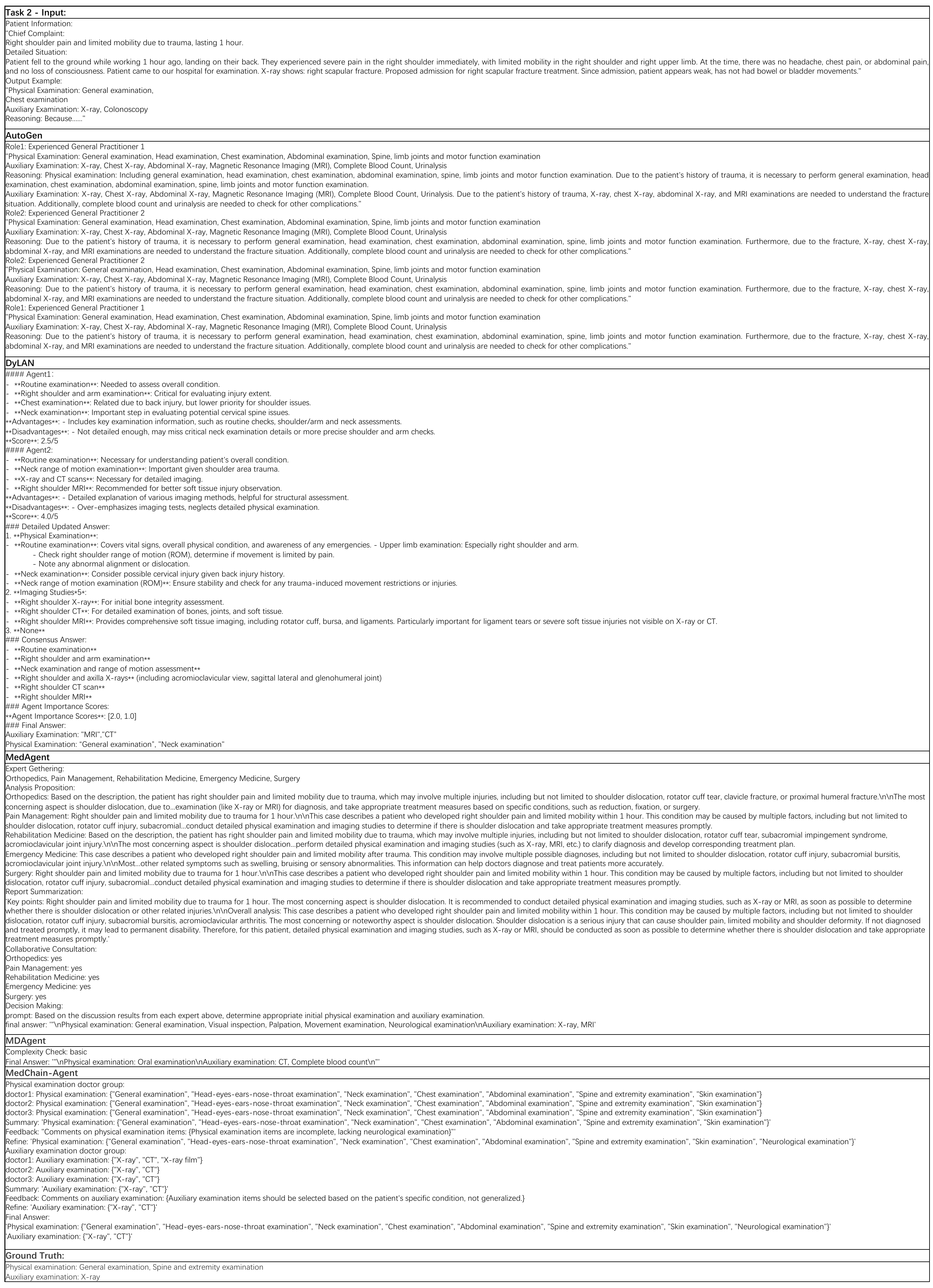}
	\caption{\textbf{The case study among different multi-agent for task2.} }
	\label{fig:case_study_2}
\end{figure*}

\begin{figure*}
	\centering
	\includegraphics[width=0.6\linewidth]{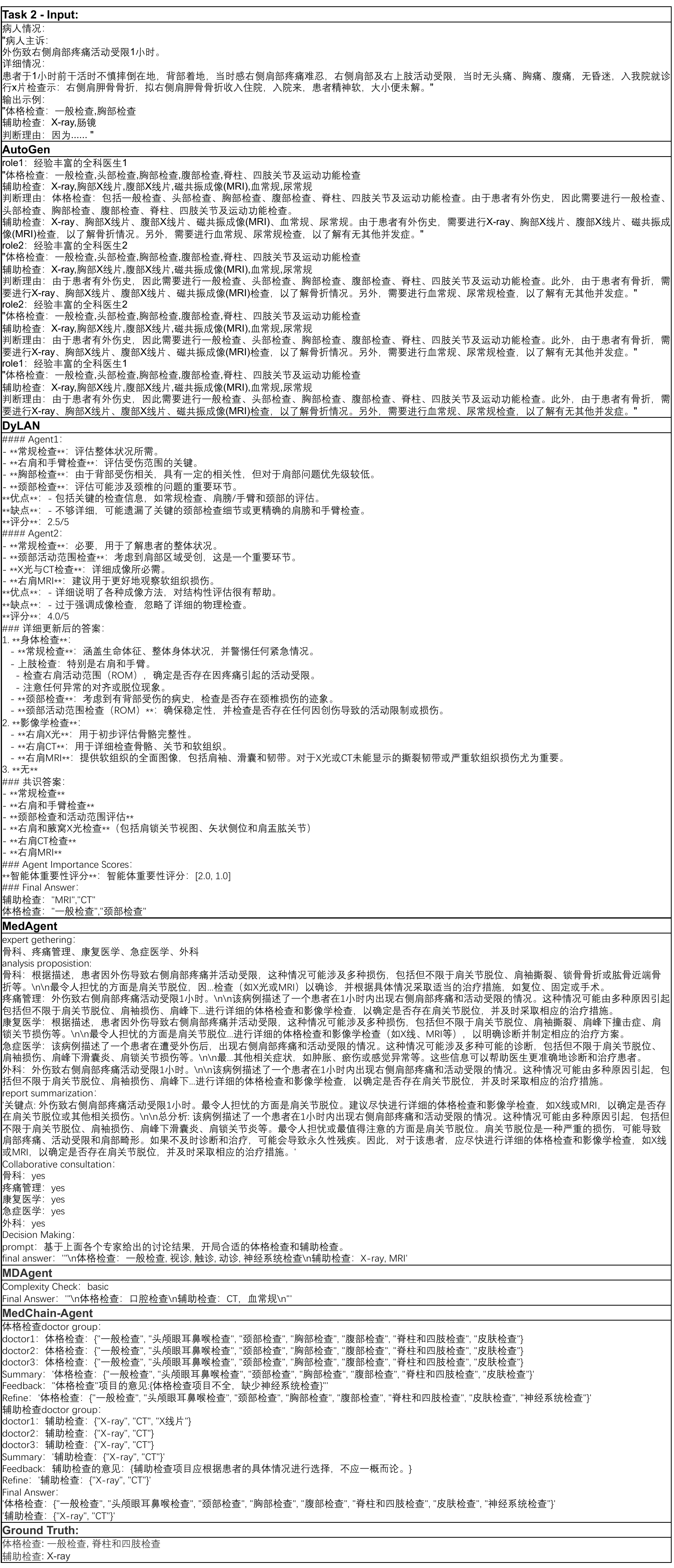}
	\caption{\textbf{The case study among different multi-agent for task2 in Chinese.} }
	\label{fig:case_study_2_chinese}
\end{figure*}

\begin{figure*}
	\centering
	\includegraphics[width=0.8\linewidth]{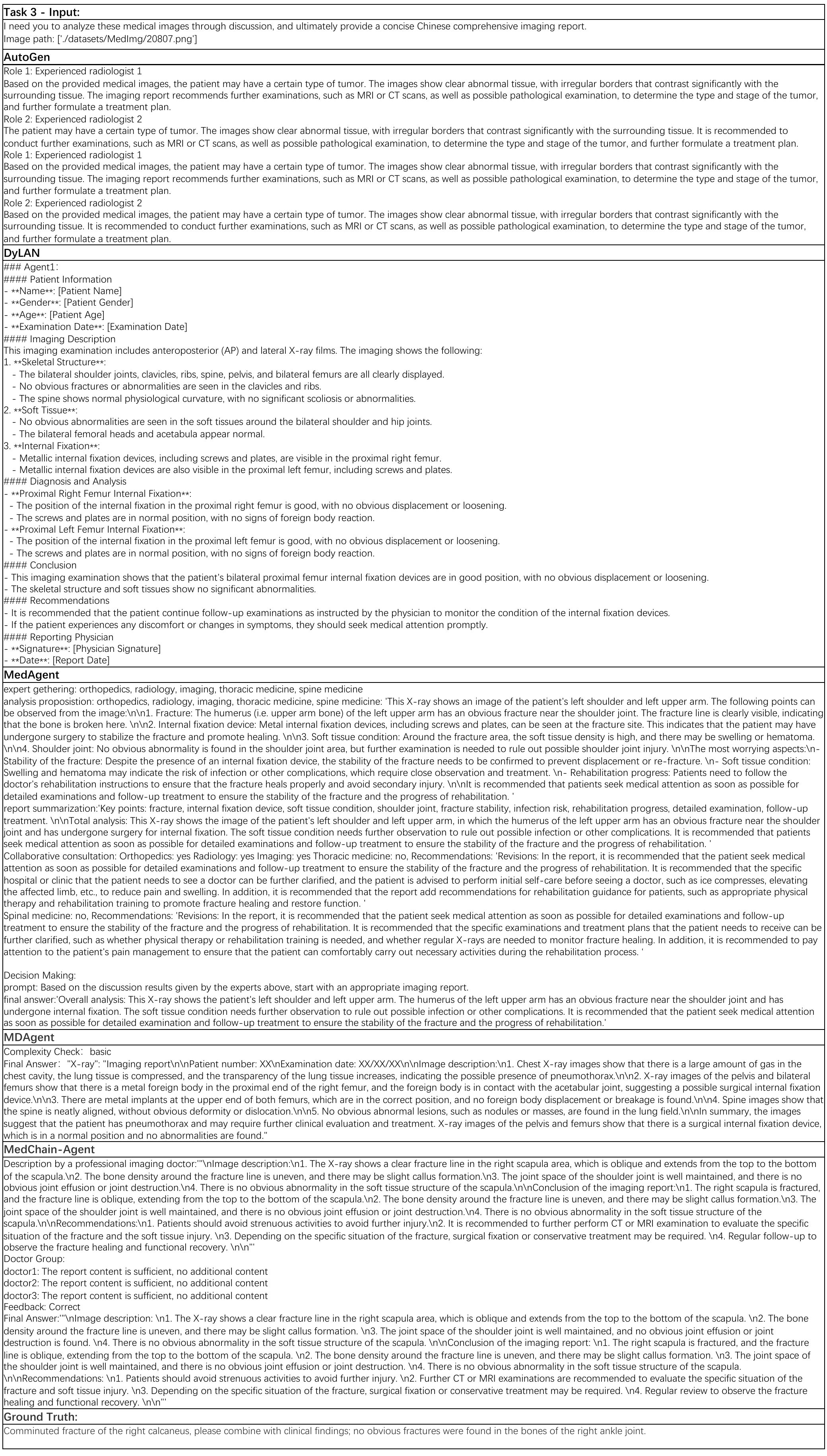}
	\caption{\textbf{The case study among different multi-agent for task3.} }
	\label{fig:case_study_3}
\end{figure*}

\begin{figure*}
	\centering
	\includegraphics[width=1\linewidth]{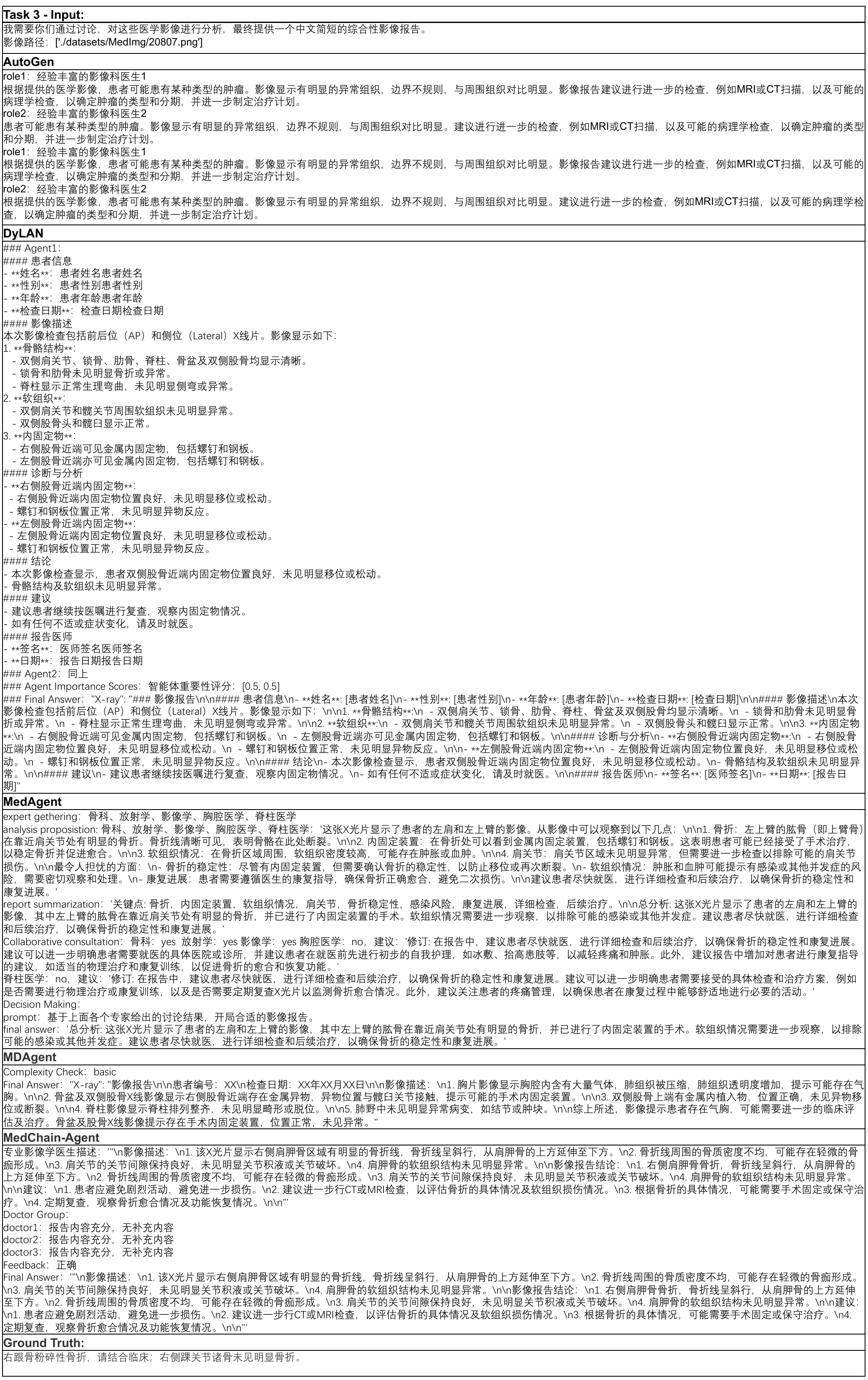}
	\caption{\textbf{The case study among different multi-agent for task3 in Chinese.} }
	\label{fig:case_study_3_chinese}
\end{figure*}

\begin{figure*}
	\centering
	\includegraphics[width=1\linewidth]{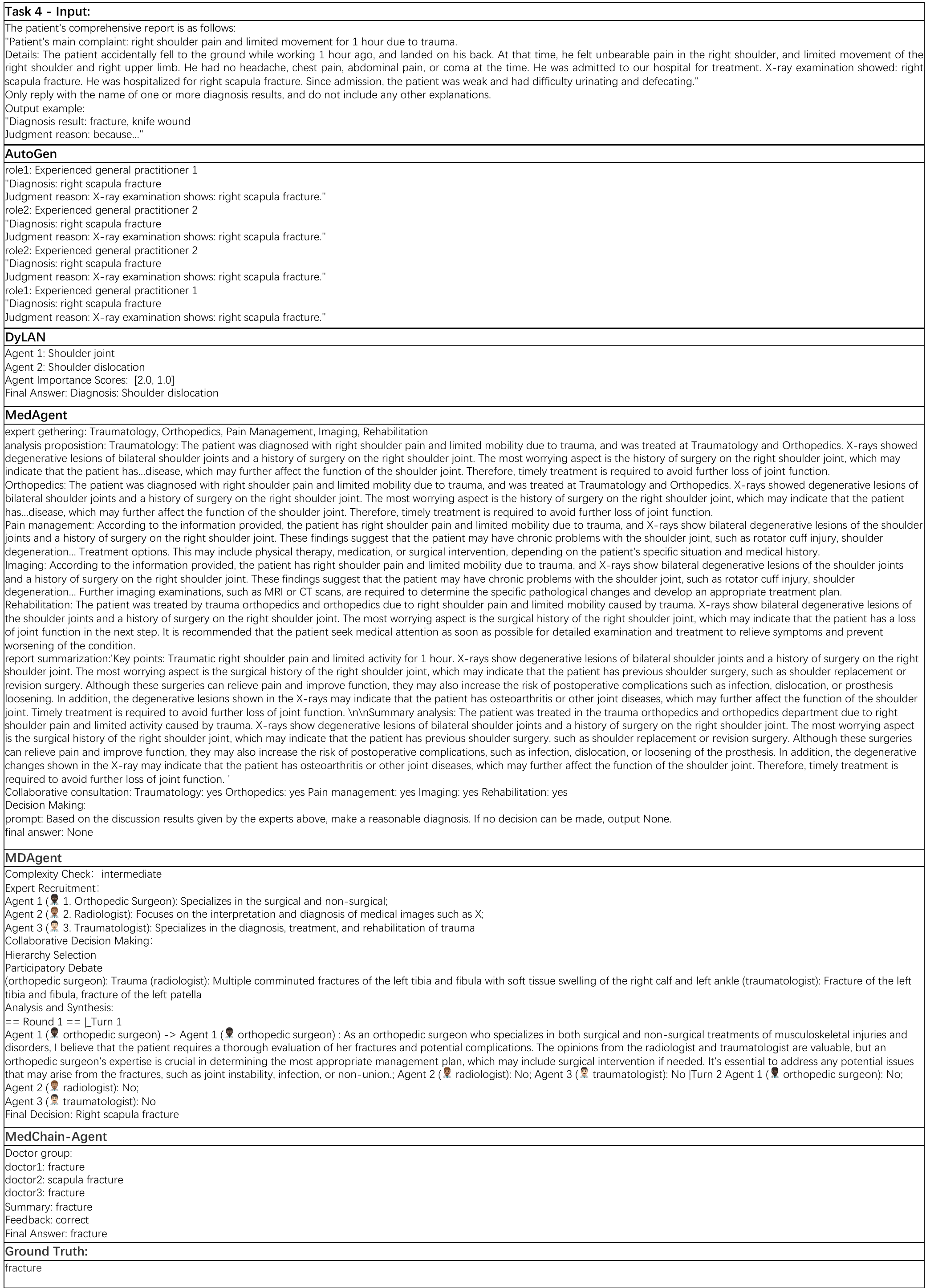}
	\caption{\textbf{The case study among different multi-agent for task4.} }
	\label{fig:case_study_4}
\end{figure*}

\begin{figure*}
	\centering
	\includegraphics[width=1\linewidth]{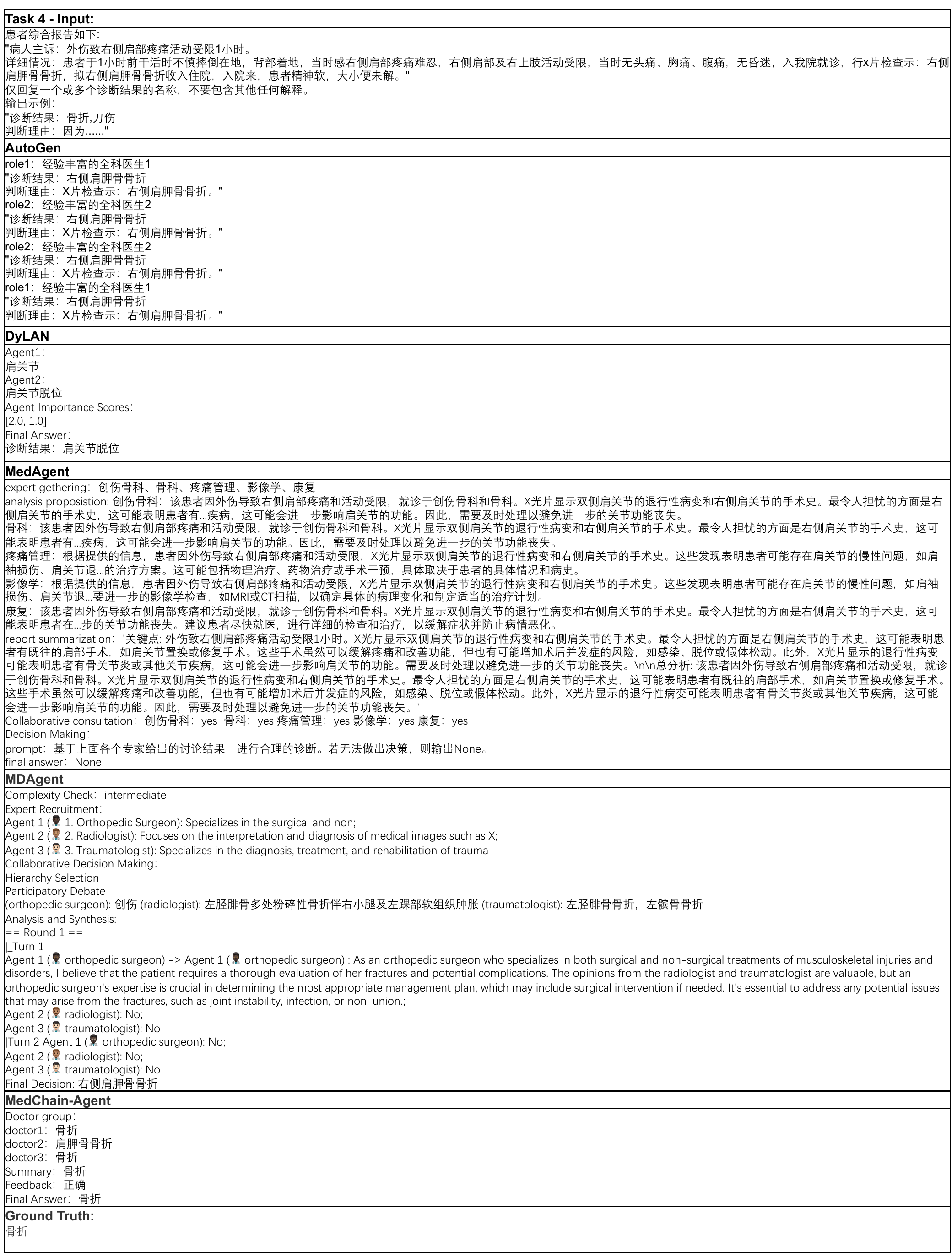}
	\caption{\textbf{The case study among different multi-agent for task4 in Chinese.} }
	\label{fig:case_study_4_chinese}
\end{figure*}

\begin{figure*}
	\centering
	\includegraphics[width=1\linewidth]{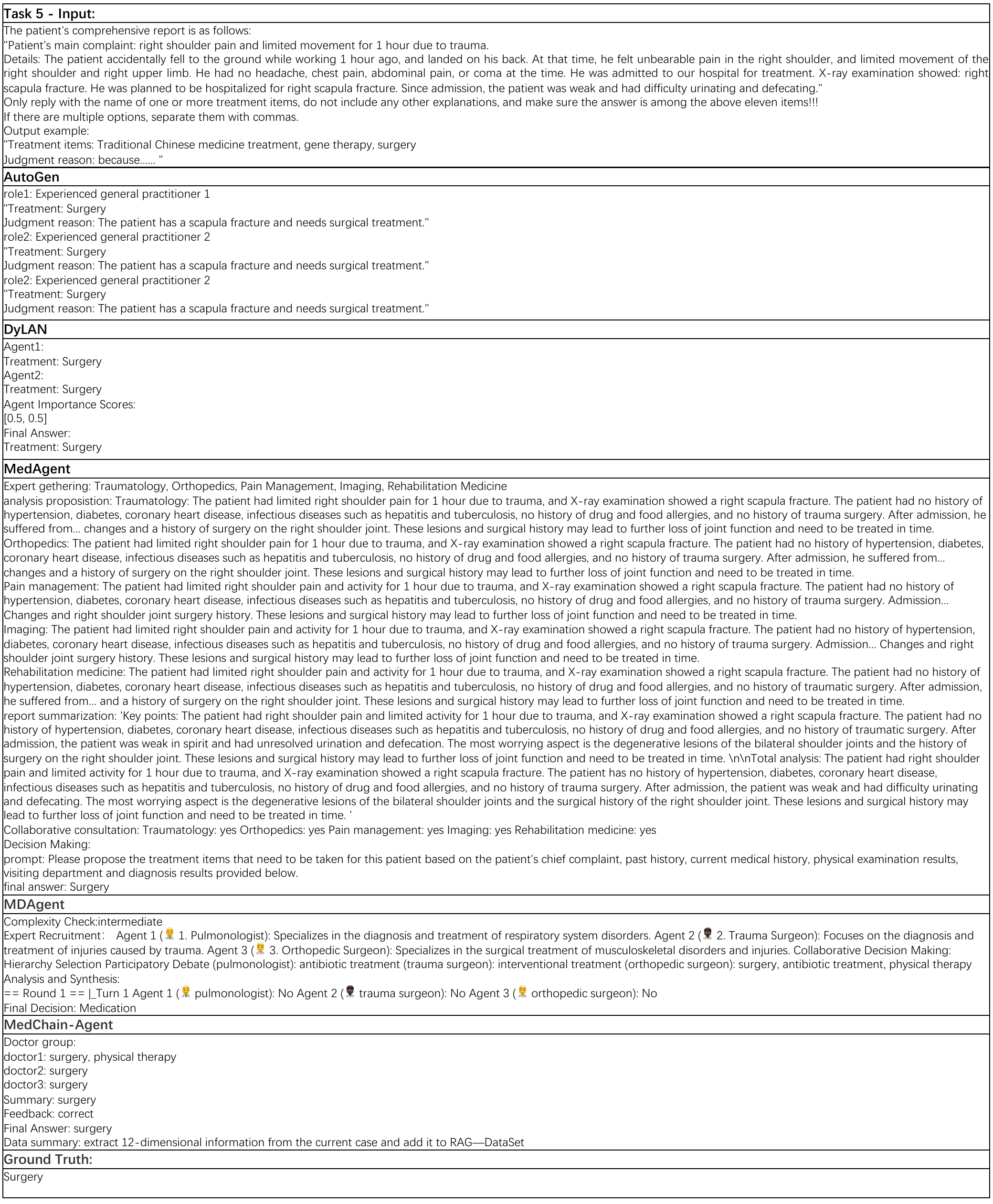}
	\caption{\textbf{The case study among different multi-agent for task5.} }
	\label{fig:case_study_5}
\end{figure*}

\begin{figure*}
	\centering
	\includegraphics[width=1\linewidth]{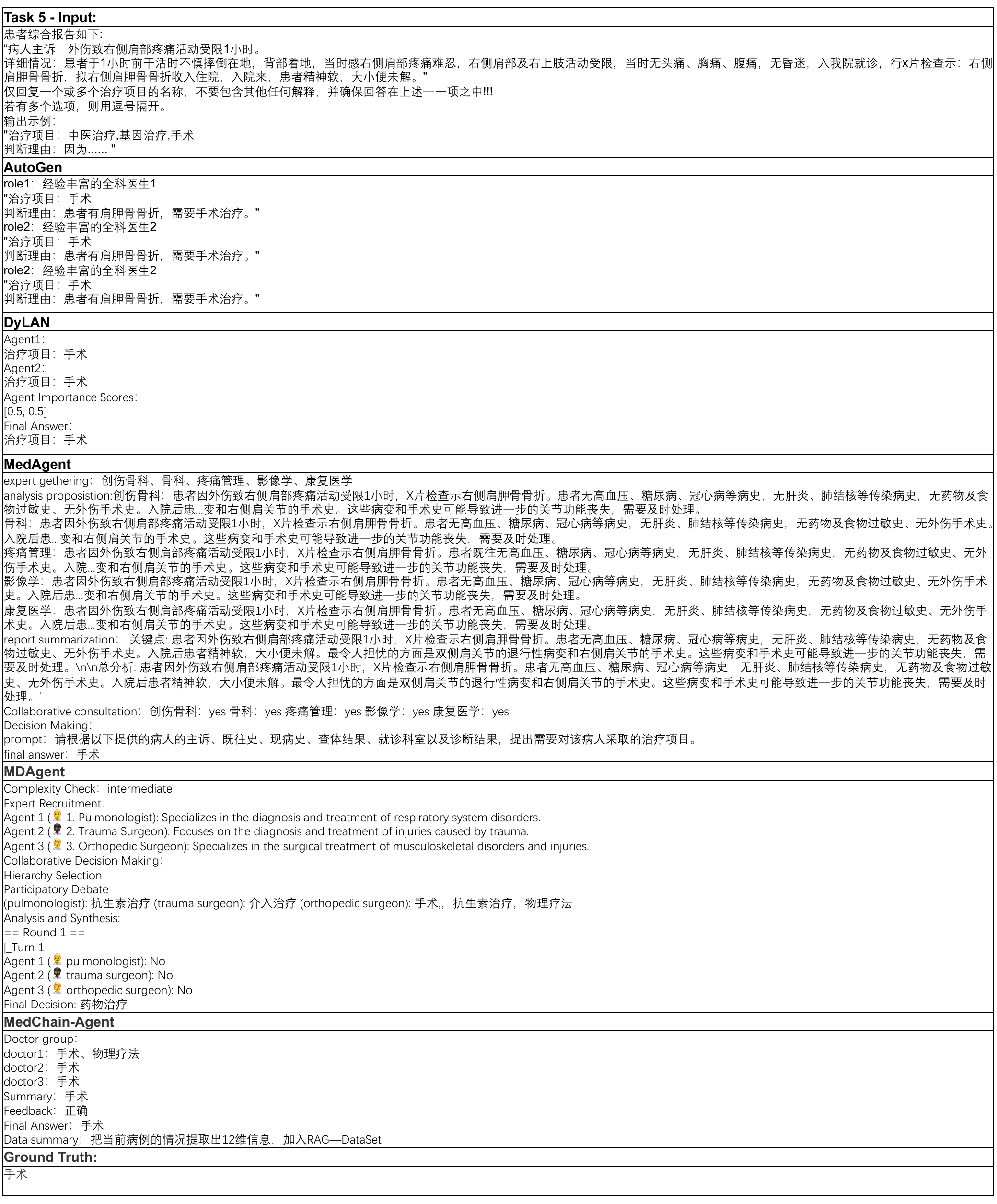}
	\caption{\textbf{The case study among different multi-agent for task5 in Chinese.} }
	\label{fig:case_study_5_chinese}
\end{figure*}

\begin{figure*}
	\centering
	\includegraphics[width=0.8\linewidth]{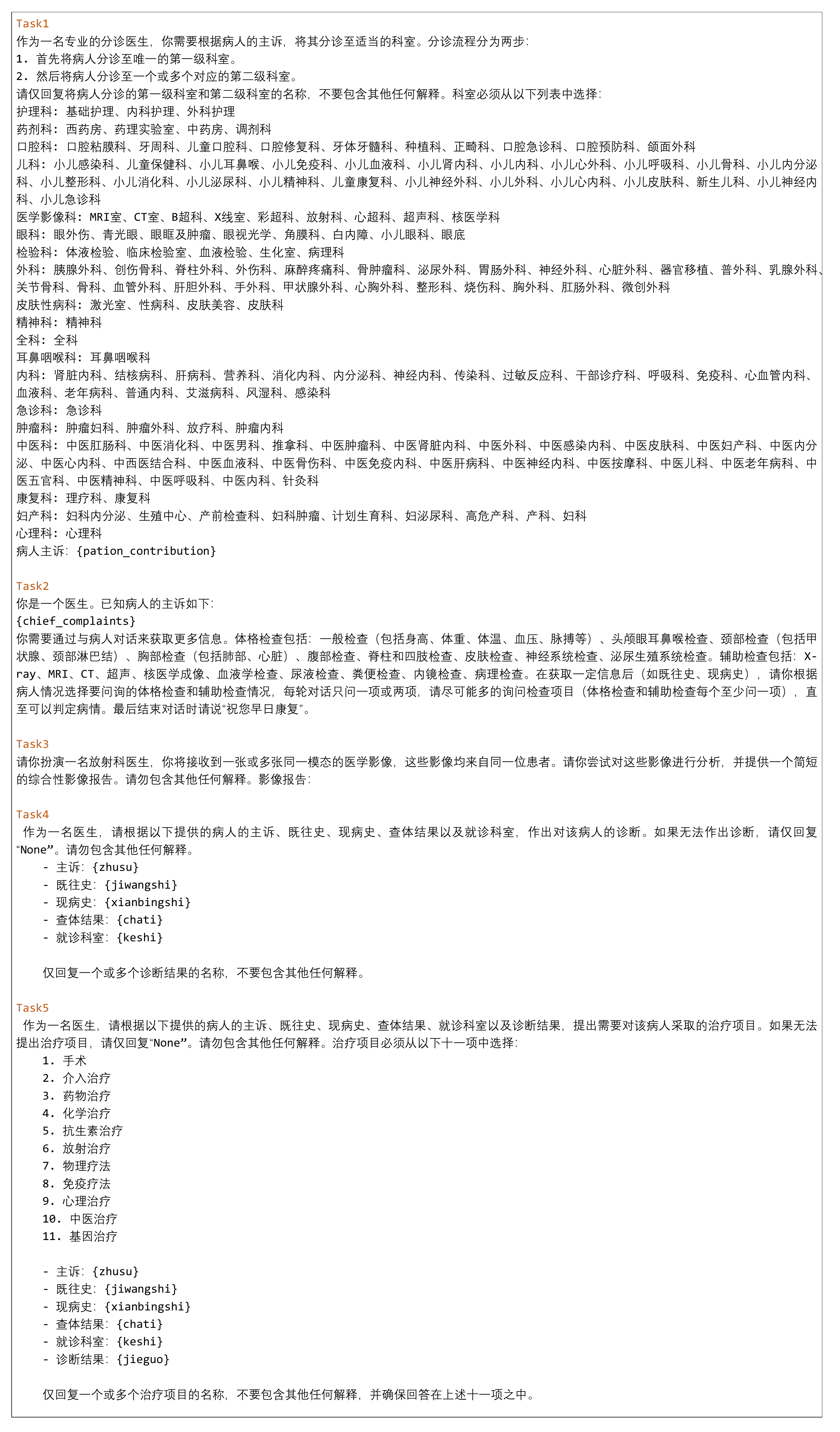}
	\caption{\textbf{The prompt of task 1 to 5 for other baseline.} }
	\label{fig:prompt_baseline}
\end{figure*}

\begin{table*}[h]
    \centering
    \caption{\textbf{Comparison of Benchmarks for LLM-based Agent.} The \benchmark composed of 12, 163 case with five sequential stages of the clinical workflow. It provide interaction environment for LLM-based agent evaluation.
    %Personal Information indicates whether the patient's basic profile is provided in the diagnostic process. Sequential Evaluation denotes whether the different stages of the diagnostic process are interconnected and build upon each other, reflecting the sequential nature of real-world clinical decision-making. Interaction signifies whether the information required for diagnosis is obtained through dynamic interaction with the patient, as opposed to being provided upfront.
    }
    \label{tab:related_works}
    \vspace{-5pt}
    \begin{small} % Reduce font size for the table
    \begin{tabularx}{\textwidth}{
        >{\raggedright\arraybackslash}p{3.5cm}
        >{\raggedright\arraybackslash}p{1.5cm}
        >{\raggedright\arraybackslash}X
        >{\centering\arraybackslash}X
        >{\centering\arraybackslash}X
        >{\centering\arraybackslash}X
    }
        \toprule
        Benchmark & \#Num & Modality & Personal Information & Sequential Evaluation & Interaction \\
        \midrule
            % VQA-RAD \cite{lau2018dataset} & 3,515 & Radiology & \no & \no & Generative QA, Multi-Choice QA \& Binary QA & Perception \& Diagnosis \\
            MedQA \cite{zhang2018medical} & 1,273 & Text & \yes & \no & \no \\
            PubMedQA \cite{jin2019pubmedqa} & 500 & Text & \no & \no & \no \\
            % MMLU (Six medical subtasks) \cite{hendrycks2020measuring} & 1,089 & Multi-Modality (no specific modalities) & \no & \no & Multi-Choice QA & Diagnosis, Pharmacotherapy, Disease Prevention, Medical Genetics, Clinical Knowledge \& Anatomy \\
            % PathVQA \cite{he2020pathvqa} & 32,799 & Radiology & \no & \no & Generative QA \& Binary QA & Perception \& Diagnosis \\
            % SLAKE \cite{liu2021slake} & 14,028 & Radiology & \no & \no & Generative QA \& Multi-Choice QA & Examination, Diagnosis \& Treatment \\
            MedMCQA \cite{pal2022medmcqa} & 193,155 & Text & \yes & \no & \no \\
            % DDXPlus \cite{fansi2022ddxplus} & 1.3 million & / & \yes & \no & Multi-Choice QA \& Binary QA & Inquiry \& Diagnosis \\
            % MedMD \cite{wu2023towards} & 16 million & Radiology & \no & \no & Generative QA \& Multi-Choice QA & Perception \& Diagnosis \\
            % PMC-VQA \cite{zhang2023pmc} & 227,000 & X-ray, MRI, CT, Ultrasound, Angiography, Echocardiography, Biopsy Imaging, Cholangiography, Fluorescence Imaging, Electron Microscopy, etc. & \no & \no & Generative QA \& Multi-Choice QA & Perception, Diagnosis, \& Patient Communication \\
            % MultiMedBench \cite{tu2024towards} & 1 million+ & CT, MRI, X-ray, Histopathology, Genomic data, Mammography & \no & \yes (7) & Generative QA \& Multi-Choice QA & Perception \& Diagnosis \\
            %ClinicalBench & 1,500 & X-ray, CT, MRI, Ultrasound & \no & \yes (24) & Generative QA \& Multi-Choice QA & Department guide, Clinical Dx \& Imaging Dx \\
            MMedBench \cite{qiu2024towards} & 53,566 & Text & \no & \no & \no \\
            MedBench \cite{liu2024medbench} & 300,901 & Text & \yes & \no & \no \\
            %AgentClinic-NEJM \cite{schmidgall2024agentclinic}& 15 & Text, Imaging & \yes & \no & \yes \\
            % AgentClinic-MedQA & 107 & / & \yes & \yes (23) & Dialogue & Dx, Exam. \& Planning \\
            % RJUA-SPs & 45 standardized patients & / &\yes & \no & Dialogue \& Generative QA & Inquiry, Exam., Dx \& Tx \\
            Asclepius \cite{wang2024asclepius} & 3,232 & Text, Imaging & \no & \no & \no\\
            GMAI-MMBench \cite{chen2024gmai} & 26,000 & Text, Imaging & \no & \no & \no\\
            \rowcolor{gray!25}
            \benchmark & 12,163 & Text, Imaging & \yes & \yes & \yes\\
        \bottomrule
    \end{tabularx}
    \end{small}
\end{table*}

\section{Supplementary for Experiment}
\label{sec:experiment_append}
Tasks 1 and 2 are subdivided based on the Chain of Thought (CoT) reasoning approach for decision-making. Consequently, Task 1 is split into "specialty referral (Level 1)" and "specialty referral (Level 2)" to evaluate performance at both primary and secondary department levels. Task 2 is divided into "Physical" and "Ancillary," representing performance in physical examinations and ancillary tests. or both ablation and comparative experiments, we split our dataset into training and test sets in a 7:3 ratio. We extracted 12-dimensional features from each training set case and stored them in our Medical Dataset for subsequent retrieval tasks.

\section{Related Work Discussion}
\label{sec:related_work}
We compare MedChain with most of related work in \autoref{tab:related_works} and \autoref{fig:benchmark_comparison}. Especially, we analyze MedChain with AI Hospital and CoD as following:

\textbf{AI Hospital \cite{fan2024ai}}: While both AI Hospital and our work aim to evaluate LLMs in clinical scenarios through multi-agent interactions, there are several key distinctions. Firstly, MedChain significantly expands the scale and diversity of medical cases, comprising 12,163 cases across 19 specialties and 156 subspecialties, including 7,338 medical images. This represents a substantial advancement over AI Hospital's dataset of 506 cases with limited specialty coverage, enabling more comprehensive evaluation of LLMs' medical capabilities across diverse clinical scenarios. Secondly, MedChain introduces a more sophisticated multi-agent collaboration mechanism. While AI Hospital employs a basic agent interaction model, our framework incorporates a feedback-driven multi-agent system enhanced by the MedCase-RAG module. This module enables dynamic knowledge base expansion and supports case-based reasoning through structured feature representation.

\textbf{CoD \cite{chen2024cod}} introduces an interactive approach to enhance the interpretability of medical diagnosis. While both CoD and our work incorporate interactive components, they differ substantially in both design objectives and implementation mechanisms. CoD primarily focuses on improving the interpretability of the diagnostic phase through confidence-driven interactions that demonstrate the reasoning process. In contrast, MedChain takes a more comprehensive approach by simulating the complete clinical workflow through a multi-agent collaborative framework. Our framework encompasses five sequential stages from triage to treatment, with each stage's decisions being evaluated and guided by subsequent stages through a novel cross-stage feedback mechanism. This design enables MedChain to capture the interdependent nature of clinical decision-making, where decisions at each stage influence and are influenced by other stages in the workflow.
%%%%%%%%%%%%%%%%%%%%%%%%%%%%%%%%%%%%%%%%%%%%%%%%%%%%%%%%%%%%

\clearpage
\newpage
\section*{NeurIPS Paper Checklist}

\begin{enumerate}

\item {\bf Claims}
    \item[] Question: Do the main claims made in the abstract and introduction accurately reflect the paper's contributions and scope?
    \item[] Answer: \answerYes{} % Replace by \answerYes{}, \answerNo{}, or \answerNA{}.
    \item[] Justification: See abstract and Section 1.
    \begin{itemize}
        \item The answer NA means that the abstract and introduction do not include the claims made in the paper.
        \item The abstract and/or introduction should clearly state the claims made, including the contributions made in the paper and important assumptions and limitations. A No or NA answer to this question will not be perceived well by the reviewers. 
        \item The claims made should match theoretical and experimental results, and reflect how much the results can be expected to generalize to other settings. 
        \item It is fine to include aspirational goals as motivation as long as it is clear that these goals are not attained by the paper. 
    \end{itemize}

\item {\bf Limitations}
    \item[] Question: Does the paper discuss the limitations of the work performed by the authors?
    \item[] Answer: \answerYes{} % Replace by \answerYes{}, \answerNo{}, or \answerNA{}.
    \item[] Justification: See section 7.

\item {\bf Theory assumptions and proofs}
    \item[] Question: For each theoretical result, does the paper provide the full set of assumptions and a complete (and correct) proof?
    \item[] Answer: \answerNA{} % Replace by \answerYes{}, \answerNo{}, or \answerNA{}.
    \item[] Justification: The paper does not include theoretical results.

    \item {\bf Experimental result reproducibility}
    \item[] Question: Does the paper fully disclose all the information needed to reproduce the main experimental results of the paper to the extent that it affects the main claims and/or conclusions of the paper (regardless of whether the code and data are provided or not)?
    \item[] Answer: \answerYes{} % Replace by \answerYes{}, \answerNo{}, or \answerNA{}.
    \item[] Justification: See section 5 and Appendix A, B, and C.

\item {\bf Open access to data and code}
    \item[] Question: Does the paper provide open access to the data and code, with sufficient instructions to faithfully reproduce the main experimental results, as described in supplemental material?
    \item[] Answer: \answerYes{} % Replace by \answerYes{}, \answerNo{}, or \answerNA{}.
    \item[] Justification: Dateset is uploaded in https://huggingface.co/datasets/ljwztc/MedChain. Code will be released in https://github.com/ljwztc/MedChain.

\item {\bf Experimental setting/details}
    \item[] Question: Does the paper specify all the training and test details (e.g., data splits, hyperparameters, how they were chosen, type of optimizer, etc.) necessary to understand the results?
    \item[] Answer: \answerYes{} % Replace by \answerYes{}, \answerNo{}, or \answerNA{}.
    \item[] Justification: See section 5 and Appendix A, B, and C.

\item {\bf Experiment statistical significance}
    \item[] Question: Does the paper report error bars suitably and correctly defined or other appropriate information about the statistical significance of the experiments?
    \item[] Answer: \answerNA{} % Replace by \answerYes{}, \answerNo{}, or \answerNA{}.
    \item[] Justification: All model weights are obtained from their official repositories on Hugging
    Face to ensure consistency and reliability.

\item {\bf Experiments compute resources}
    \item[] Question: For each experiment, does the paper provide sufficient information on the computer resources (type of compute workers, memory, time of execution) needed to reproduce the experiments?
    \item[] Answer: \answerYes{} % Replace by \answerYes{}, \answerNo{}, or \answerNA{}.
    \item[] Justification: See section 5.1.
    
\item {\bf Code of ethics}
    \item[] Question: Does the research conducted in the paper conform, in every respect, with the NeurIPS Code of Ethics \url{https://neurips.cc/public/EthicsGuidelines}?
    \item[] Answer: \answerYes{} % Replace by \answerYes{}, \answerNo{}, or \answerNA{}.
    \item[] Justification: The research fully conforms to the NeurIPS Code of Ethics, adhering to all
    ethical guidelines without deviation.

\item {\bf Broader impacts}
    \item[] Question: Does the paper discuss both potential positive societal impacts and negative societal impacts of the work performed?
    \item[] Answer: \answerYes{} % Replace by \answerYes{}, \answerNo{}, or \answerNA{}.
    \item[] Justification: See in section 6.
    
\item {\bf Safeguards}
    \item[] Question: Does the paper describe safeguards that have been put in place for responsible release of data or models that have a high risk for misuse (e.g., pretrained language models, image generators, or scraped datasets)?
    \item[] Answer: \answerNA{} % Replace by \answerYes{}, \answerNo{}, or \answerNA{}.
    \item[] Justification: Not applicable.

\item {\bf Licenses for existing assets}
    \item[] Question: Are the creators or original owners of assets (e.g., code, data, models), used in the paper, properly credited and are the license and terms of use explicitly mentioned and properly respected?
    \item[] Answer: \answerYes{} % Replace by \answerYes{}, \answerNo{}, or \answerNA{}.
    \item[] Justification: We obtain the permission from the website.

\item {\bf New assets}
    \item[] Question: Are new assets introduced in the paper well documented and is the documentation provided alongside the assets?
    \item[] Answer: \answerYes{} % Replace by \answerYes{}, \answerNo{}, or \answerNA{}.
    \item[] Justification: See the abstract for a link to the dataset, website, and the code, which include
    details about our new benchmark

\item {\bf Crowdsourcing and research with human subjects}
    \item[] Question: For crowdsourcing experiments and research with human subjects, does the paper include the full text of instructions given to participants and screenshots, if applicable, as well as details about compensation (if any)? 
    \item[] Answer: \answerNo{} % Replace by \answerYes{}, \answerNo{}, or \answerNA{}.
    \item[] Justification: Not applicable.

\item {\bf Institutional review board (IRB) approvals or equivalent for research with human subjects}
    \item[] Question: Does the paper describe potential risks incurred by study participants, whether such risks were disclosed to the subjects, and whether Institutional Review Board (IRB) approvals (or an equivalent approval/review based on the requirements of your country or institution) were obtained?
    \item[] Answer: \answerNo{} % Replace by \answerYes{}, \answerNo{}, or \answerNA{}.
    \item[] Justification: Not applicable.

\item {\bf Declaration of LLM usage}
    \item[] Question: Does the paper describe the usage of LLMs if it is an important, original, or non-standard component of the core methods in this research? Note that if the LLM is used only for writing, editing, or formatting purposes and does not impact the core methodology, scientific rigorousness, or originality of the research, declaration is not required.
    %this research? 
    \item[] Answer: \answerYes{} % Replace by \answerYes{}, \answerNo{}, or \answerNA{}.
    \item[] Justification: We test the performance of LLM in our dataset.

\end{enumerate}

\end{document}